\documentclass[11pt]{article}
\usepackage{blindtext, graphicx}
\usepackage{listings}
\usepackage[left=1in,top=1in,right=1in,bottom=1in,nohead]{geometry}
\makeatletter
\def\ps@pprintTitle{%
 \let\@oddhead\@empty
 \let\@evenhead\@empty
 \let\@oddfoot\@empty
 \let\@evenfoot\@empty
}
\usepackage{graphicx}
\usepackage[font=small]{caption}
\usepackage{amsmath,amsfonts,amssymb}

\usepackage{multirow, multicol}
\usepackage[table,xcdraw]{xcolor}
\usepackage{tabularx,booktabs, longtable, supertabular}
\usepackage{booktabs,xltabular}
\usepackage{color}
\definecolor{DarkGreen}{rgb}{0.2,0.5,0.2} 
\usepackage{hyperref}
\hypersetup{colorlinks,linkcolor={blue},citecolor={blue},urlcolor={red}}  
\usepackage[numbers]{natbib}

\usepackage{algorithm}
\usepackage{algpseudocode}

\usepackage[utf8]{inputenc}
\usepackage{longtable}
\usepackage{authblk}

\title{PDx---Adaptive Credit Risk Forecasting Model in Digital Lending using Machine Learning Operations}
\author[1]{Sultan Amed}
\author[2]{Chan Yu Hang}
\author[1]{Sayantan Banerjee}

\affil[1]{OM \& QT Area, Indian Institute of Management Indore}

\affil[2]{Department of Statistics \& Data Science, National University of Singapore}

\date{}

\begin{document}
\maketitle

\begin{abstract}
This paper presents PDx, an adaptive, machine learning operations (MLOps) driven decision system for forecasting credit risk using probability of default (PD) modeling in digital lending. While conventional PD models prioritize predictive accuracy during model development with complex machine learning algorithms, they often overlook continuous adaptation to changing borrower behaviour, resulting in static models that degrade over time in production and generate inaccurate default predictions. Many financial institutes also find it difficult transitioning ML models from development environment to production and maintaining their health. With PDx we aimed to addresses these limitations using a dynamic, end-to-end model lifecycle management approach that integrates continuous model monitoring, retraining, and validation through a robust MLOps pipeline. We introduced a dynamic champion-challenger framework for PDx to regularly update baseline models to recalibrate independent parameters with the latest data and select the best-performing model through out-of-time validation, ensuring resilience against data drift and changing credit risk patterns. Our empirical analysis shows that decision tree-based ensemble models consistently outperform others in classifying defaulters but require frequent updates to sustain performance. Linear models (e.g., logistic regression) and neural networks exhibit greater performance degradation. The study demonstrate with PDx we can mitigates value erosion for digital lenders, particularly in short-term, small-ticket loans, where borrower behavior shifts rapidly. We have validated the effectiveness of PDx using datasets from peer-to-peer lending, business loans, and auto loans, demonstrating its scalability and adaptability for modern credit risk forecasting.

\end{abstract}



\section{Introduction}
Advancements in technology and the rise of cost-effective platforms have revolutionized digital lending \cite{cornelli2023fintech}. The proliferation of digital lending has played an essential role in fostering financial inclusion and expanding the accessibility of lending services to a broader spectrum of customers \cite{bazarbash2020filling}. Between 2011 and 2021, the World Bank's global financial index reported a noteworthy growth in financial inclusion, rising from 56 to 80 percent across gender, income groups, and educational backgrounds \cite{demirgucc2022global}. Despite this progress, the increased outreach to unbanked customers has led to a surge in default events, highlighting the weaknesses in existing credit assessment methodologies. This trend has notably impacted the interests of lenders and investors as misclassifying defaults results in substantial economic losses \cite{xia2017cost}. As these digital lending platforms ascend in popularity and significance, the imperative for developing advanced Probability of Default (PD) models becomes evident. These models are essential for upholding financial stability and sustainability in the evolving landscape of lending practices. 

In the realm of digital lending, credit risk models serve as automated tools crucial for appraising loan applications, aiming to ascertain approval or denial. Financial institutions across the globe use diverse credit risk measures to manage their consumer asset loans effectively. Among these metrics, the most critical one is Expected Loss (EL), calculated as the product of PD, representing the probability of default, EAD denoting exposure at default, and LGD, the loss given default \cite{papouskova2019two}. These models categorize borrowers into different risk buckets based on their creditworthiness, utilizing data gathered during the application phase. This aids lenders in critical loan decisions, including approvals, pricing, and offering appropriate credit limits \cite{thomas2017credit}. However, the digital era has brought constant changes in the credit landscape, influenced by various factors such as economic shifts, policy changes, evolving borrower behaviors, product innovation, data development, and technological progress. These factors cause the incoming data subject to continuous change, leading to inaccuracies and reduced effectiveness of PD models over time. Current research primarily focuses on optimizing algorithms for credit scoring, but often overlooks the crucial stages following model finalization and deployment.
Additionally, there is a lack of practical consideration for time-lag issues in analytics models across development, testing, validation, and production stages. For instance, in PD modeling, a common target variable is `90DPD in 12 months post-loan disbursal,' meaning that by the time the model is deployed, it's already based on data that's 12 months old, even if the latest data was used for training.  There is also a notable gap in the comprehensive understanding and management of the PD model lifecycle. As Sculley et al. \cite{sculley2015hidden} highlight, the architecture of real-world machine learning (ML) systems for credit scoring extends beyond just the machine learning code. This highlights the importance of considering the broader components of these models to fully grasp their role and impact in the credit assessment process.

A model is a function of data and algorithm, \emph{f(data, algorithm)}. As the data evolves, it is crucial to retest the underlying algorithm with the new data and compare its performance against other promising algorithms, ensuring no opportunities for performance enhancement are missed. Traditional PD model research, often based on static data and focusing on a few parts of the model lifecycle, tends to result in inaccuracies and inefficiencies, leading to failures in production. Machine Learning Operations (MLOps) \cite{kreuzberger2023machine}, an approach drawing from best practices in software engineering and initially applied in fields like image classification \cite{testi2022mlops}, offers a solution to these challenges by facilitating a comprehensive end-to-end management of the PD model lifecycle. MLOps enables the automation of various stages, from data collection and preprocessing to model training, evaluation, deployment, and ongoing monitoring. This automation not only streamlines the entire process but also minimizes errors and continuously improves the PD models. By integrating MLOps, lenders can access more accurate and current credit risk assessments, enhancing responsible and inclusive lending practices. The main contributions of our work are outlined as follows: 

\renewcommand{\labelenumi}{\alph{enumi})}
\begin{enumerate}
    \item A comprehensive literature review identifying the significant research gaps in the development of existing credit risk forecasting (PD scoring) models.
    \item A novel PDx model framework for credit risk forecasting, employing a continuous champion vs. challenger online modeling approach in production, enhanced and updated regularly using MLOps CICD pipelines.
    \item Empirical evaluations of the proposed method in three credit datasets, utilizing multiple performance metrics to stress-test performance over existing benchmark methods, in the out-of-time production window.
    \item A proposed cloud-based model serving framework to deploy PDx and govern its life-cycle, ensuring reproducibility of results.

\end{enumerate}

The remainder of the paper is structured into six sections: Section~\ref{sec:lit-review} reviews the use of AI / ML in credit risk forecasting research. Section~\ref{sec:methods} outlines our methodology, and Section~\ref{sec:expt-design} describes the experiment design and dataset. Section~\ref{sec:results} presents a performance evaluation of our model against existing methods. Section~\ref{sec:discussion} discusses the findings and implications. The paper concludes in Section~\ref{sec:conclusion} with a summary, a discussion of limitations, and future research directions in credit scoring and MLOps.


\section{Literature Review}
\label{sec:lit-review}

The core principle of credit assessment is to compare a customer's profile with those of past borrowers who successfully repaid their loans. Financial institutions are more likely to approve credit for customers whose profiles resemble these reliable borrowers \cite{abdou2011credit}. Traditionally, this evaluation involved two approaches: subjective assessments by loan officers and objective credit scoring methods \cite{desai1996comparison}. While subjective assessment was once predominant, the digital era has shifted towards machine learning-based credit scoring. This automated approach is now the primary method for efficient loan decision-making, with most applications processed digitally and only a few flagged for manual review by loan officers or field verification. This shift is driven by the superior performance of machine algorithms on large-scale digital lending decisions, leading to cost-effective loan approval processes \cite{jansen2023rise}. This has sparked significant research interest in advancing credit assessment models in academic and industry circles.

As summarized in Table \ref{table:previous-studies}, extensive research has been conducted on developing PD models for credit scoring. These investigations span various aspects, including the creation of optimal classification algorithms, identification of relevant data sources and drivers, feature engineering techniques, selection of appropriate performance metrics, and comparisons of different models to determine the most suitable algorithm for given data \citep{papouskova2019two,baesens2023boosting,louzada2016classification,abellan2017comparative,dastile2020statistical}. While it started with linear models that are easy to interpret, such as logistic regression, LDA, weighted scorecards \citep{wiginton1980note,hand2002superscorecards}, etc., recent trends suggest that the focus of many researchers in the field of credit risk forecasting has shifted towards exploring and developing models based on machine learning and deep learning algorithms \cite{zhang2019online}. More complex algorithms have been employed to train the PD model, aiming to enhance predictive performance, even for marginal gains measured in basis points (bps). A few noteworthy algorithms that are exploited heavily include Decision Trees, Maximum Margin classifiers, e.g., Support Vector Machine (SVM) \cite{maldonado2017integrated}, an ensemble of bagging trees e.g., Random Forest (RF) \citep{arora2020bolasso,dastile2020statistical}, an ensemble of boosted trees e.g., Adaboost, GBM, XGBoost, \cite{gunnarsson2021deep} etc. deep learning, e.g., NN-MLP, \citep{xia2022deep,fu2020listening,marcano2011artificial}, LSTM \cite{shen2021new},  etc., and complex ensemble of multiple algorithms such as stacking \cite{papouskova2019two}. While a lot of focus has been given to training the PD model using complex neural networks,  Gunnarsson et al. \cite{gunnarsson2021deep} in their benchmark study with ten different type credit datasets suggest that XGBoost and RF outperform other algorithms when it comes to choosing a base algorithm for developing PD model. This finding is echoed by Grinsztajn et al. \cite{grinsztajn2022tree}, who noted that despite deep learning's progress in other domains, tree-based models like XGBoost and Random Forests tend to be more effective for tabular data. To enhance the performance of PD models, significant efforts have been made in developing and optimizing new algorithms and exploring various data sources \cite{baesens2023boosting} and feature engineering techniques that provide incremental insights into credit default behavior and boost predictive accuracy. Djeundje et al. \cite{djeundje2021enhancing} demonstrated adding psychometric and alternative data over primary demographic data can significantly improve model performance. Zhang et al. \cite{zhang2020credit} derived textual features from loan descriptions to improve overall model performance for P2P lending institutes. Óskarsdóttir et al. \cite{oskarsdottir2019value} demonstrated that incorporating alternative data along with better modeling techniques in the credit scoring process could lead to superior model performance. Traditionally, credit scoring has been viewed as a classification problem; however, Bellotti et al. \cite{bellotti2009credit} and Glennon et al. \cite{glennon2005measuring} introduced the concept of time-to-default, treating credit scoring as a survival event, and Brezigar et al. \cite{brezigar2021modeling} employed probit and tobit modeling techniques to predict days past due (DPD), thereby enhancing model performance with alternative objective functions. Addressing low default events, several authors also explored PD as a class imbalance classification problem. They suggested different oversampling, e.g., SMOTE and under-sampling techniques, to handle it and further improve model performance \citep{papouskova2019two, liu2023tree, song2023loan}.


\begin{table*}[h]
\centering
\caption[]{Summary of past research on Credit Scoring Model}
\label{table:previous-studies}
\addtocounter{footnote}{3}
\resizebox{\textwidth}{!}
{
 \begin{tabular}{|c|c|c|c|c|c|l|}
\hline
  Domain &   Publication &   Modeling &   Key Concept &   Performance  & Validation & No. of\\
  
  & & Technique \footnotemark[1] & addressed & Metrics \footnotemark[2]  & approach \footnotemark[3] & dataset\\ \hline

  Data &
  \cite{stevenson2021value} &   LR, RF, DL &   Text data mining, Process automation &  AUC, Brier Score & ITV, OTV & 1 \\

  & \cite{djeundje2021enhancing} &   XGBoost &  Alternative Data, Predictive performance &   AUC & ITV & 4\\
  
  & \citep{oskarsdottir2019value,gambacorta2019machine} & LR, DT, RF &  Alternative Data, FE, Model performance &   AUCROC & ITV, OTV & 2, 1\\
 
  &   \cite{fernandes2016spatial} &  LR &  Alternative Data, Feature Engineering &  KS, Gini & ITV & 1\\
  
  & \cite{bellotti2009credit} &  CoX PH, LR &  Macroeconomic variable, Survival modeling &
  Mean cost & OTV & 1\\

  &  \cite{vsuvstervsivc2009consumer} &  PCA, DT & Low data scenario &
  Accuracy & ITV & 1\\

  & \cite{boughaci2018new} & Bayesian Network & Variable selection &  Precision, Recall, F-Score & 10-fold CV (ITV) & 4\\ 

  & \cite{koutanaei2015hybrid} &  PCA, GA, ET (Ensamble) &  Feature Engineering and selection &  Accuracy, AUC  & ITV & 1\\

  & \cite{brezigar2021modeling} &  Probit, Tobit &  New Target Variable (DPD modeling) &
  Accuracy, Error Matrix & ITV & 1\\

  & \cite{glennon2005measuring} &  DTHM & New target variable (Time to event) &  Hazard rate & DEV, ITV & 1\\ 
    
  &    &    &    &  &  & \\


  Model & \cite{he2023privacy} &  FL &  Data Privacy, Decentralize ML &
  Accuracy, Error Matrix & ITV & 1\\

  & \citep{xia2022deep,fu2020listening,marcano2011artificial} & NN &   Predictive performance &  AUC & 80/20,5-fold,ITV & 4,1,2\\ 

  & \citep{liu2023tree,song2023loan,finlay2011multiple} & ET & Class Imbalance, predictive performance & Accuracy, AUC, F-Score & ITV & 4, 1, 2\\

  & \citep{lextrait2023scaling,gunnarsson2021deep} &  Boosting &  Boosting predictive performance, Profit & ACC, AUC, BS, EMP(Profit) & ITV & 1, 10 \\

  & \citep{dastile2020statistical,arora2020bolasso} &  RF &  Predictive performance, Feature selection &  AUC, BS, EMP, & ITV & 2, 3\\

  & \citep{zhou2010least,maldonado2017integrated} &  SVM &  Profit maximization, predictive performance &  AUC, Acc, Profit & ITV & 2, 2\\

  & \citep{medina2023joint,bai2022gradient} &  Survival (SA) &  Time to default forecasting &  AUC, Hazard Rate, BS & ITV & 1, 2\\

  &  \citep{simumba2022multiple,kozodoi2019multi} &  GA &  Profit Scoring, Multi-objective Optimization &  EMP, No. of features & ITV & 1, 10\\ 
  
  & \cite{hoffmann2007inferring} &  FR, EA &  Modeling approach, predictive performance &   Accuracy, Error Matrix & ITV & 7\\

  & \cite{hand2002superscorecards} &  LR, Superscorecard & Improved decisioning &  Capture Rate (Bad) & DEV, ITV & 1\\

  & \cite{wiginton1980note} &  LR &  Predictive power, LR vs LDA &
  Accuracy & ITV & 1 \\

  &    &    &    &    &   &
   \\


  Interpretability &   \cite{chen2024interpretable} &  XGBoost, LIME, SHAP &   Explainable AI for model interpretation &  Variable \& co-eff. stability index & ITV & 1\\

  and Fairness &  \cite{zhu2023explainable} & Boosting, DT, LR, LIME & Local interpretation of determinants &  AUC, Accuracy, Precision, PSI & ITV & 1\\

  & \cite{sachan2020explainable} &   Belief-Rule-Base (BRB) &   Explainable AI for model interpretation &   Precision, Recall, F-Score & ITV & 1\\

  & \cite{ariza2020explainability} &  LR,DT, RF, XGB, SHAP & Accurate and transparent ML models & ACC, AUC, KS, F-Score & DEV,ITV & 1\\

  &  \cite{moldovan2023algorithmic} &  Bias mitigation methods &  Fair AI Credit Scoring & DI, SP, AOD, EOD, TI, BAcc & ITV & 2\\

  & \citep{kozodoi2022fairness,jammalamadaka2023responsible} &   Fair ML, XGBoost &
  Responsible and explainable credit scoring,  & AUC, F1-score, DI, SP, Profit & ITV & 7, 1\\

  &    &    &    &    &  & \\


  Credit & \cite{kazemi2023estimation} &  GA - NN  &  Automatic identification of PD cut-offs &   EMC, AUC  & ITV & 2\\

  Decisions & \cite{herasymovych2019using} &  RL &   Credit score acceptance threshold optimization &   Profit difference & OTV & 1\\
  
  & \cite{papouskova2019two} &   ET, Stacking with RF &  Expected loss calculation &  ACC, AUC, MC, Profit & ITV & 2\\

  & \cite{shen2020three} & LR,ANN,RF,XGB, TL &  Three-stage reject inference learning  &   AUC, KS & DEV,ITV & 1\\

  & \cite{banasik2003sample} &   BPM &   Sample selection bias, Reject inference &   AUCROC & ITV (Hold out) & 1\\
 
  &    &    &    &   &  & \\


  Deployment & \cite{nikolaidis2017exploring} &  LR, Linear SVM &  Exploration of population drift &  AUC &  OTV & 1\\

  and &  \cite{zhang2019online} &  Ensemble classifier &  Handling of concept drift and class imbalance &  ACC, Recall, F-Score, G-mean, PAUC & ITV,OTV & 1\\

  Governance &  \cite{bijak2011kalman} &  Kalman filtering & Scorecard monitoring &   Capture Rate, Gini, KS & OTV & 1\\

  & \cite{medema2009practical} & NA & Validating credit risk modelling exercise & Actual vs predicted probability & ITV, OTV & 1\\ 
  &    &    &    &   &  &\\ \hline
 
  &    &    &    &   &  & \\

  E2E Credit &  \multirow{3}{*}{\textcolor{blue}{This Paper}} &   PDx - Champion/Challenger, &  Boosted credit scoring in production with &   AUC, KS &  ITV, OTV, & ~\\
  
  Scoring Pipeline &   &   Neural Network, XGBoost, &  MLOPs using CI/CD to proactively &
  Accuracy, F1-Score, Bad Rate & Production & 3\\ 

  with MLOPs &    &   RF, Regularised LR &   upgrading model and managing risk &    & validation & ~\\ \hline 
 \end{tabular}
}

\end{table*}

\footnotetext[1]{Model Abbreviation: LR -  Logictoc Regression,  DT - Decision Tree, RF - Random Forest, XGBoost - Extreme Gradient Boosting, DL - Deep Learning, NN - Neural Network, BN - Bayesian Network,  Cox PH - Cox Proportional Hazard Model, PCA - Principal Component Analysis, GA - Genetic Algorithm, ET - Ensemble Techniques, DTHM - Discrete-Time Hazard Model, SVM - Support Vector Machine, SA - Survival Analysis, EA - Evolutionary Algorithm, LIME - Local Interpretable Model-agnostic Explanations, SHAP - SHapley Additive exPlanations, BRB - Belief Rule Base,  FL - Federated Learning, RL - Reinforcement Learning, TL - Transfer Learning, BPM - Bivariate Probit Model, KL - Kalman Filter}

\footnotetext[2]{Performance Metrics Abbreviation: ACC - Accuracy, AUC - Area Under the Curve, KS - Kolmogorov–Smirnov Statistics, TPR - True Positive Rate, FPR - False Positive Rate, EMP - Expected Maximum Profit, BS - Brier Score, HR - Hazard Rate, PSI - Population Stability Index}

\footnotetext[3]{Validation approach Abbreviation: ITV: In Time Validation, K-Fold: K-fold Cross-validation, OTV - Out of Time Validation }


The development of PD or credit scoring models typically follows a structured process: a) data collection and preprocessing, b) model training, c) model validation and evaluation, d) deployment, and e) monitoring \cite{baesens2023boosting}. Recent research has evolved to address the increasing complexity and regulatory demands in model development, focusing now on key evaluation aspects. This includes enhancing model interpretability \citep{chen2024interpretable, ariza2020explainability} for more precise understanding of variables and outcomes, ensuring fairness in lending practices through fairness assessment \citep{moldovan2023algorithmic,kozodoi2022fairness}, and evaluating the business impact \cite{herasymovych2019using} for more effective credit decision-making. Post-development aspects like performance degradation over time, interpretability, fairness in production, and model failure due to data drift are areas that need further exploration. Further, many studies, such as those in Table \ref{table:previous-studies}, validate selected PD model performance using the same dataset (development sample) or an out-of-sample test set from the same period (ITV), often employing train/test splits or k-fold cross-validation. Medema et al. \cite{medema2009practical} pointed out that this approach could overestimate future accuracy, leading to quick failure of the model in production due to performance degradation. Additionally, current research often overlooks how time-induced heterogeneity in data can affect a model's performance, interpretation, and fairness over time.

\section{Methods}
\label{sec:methods}

Our proposed approach is baseline model-agnostic which offers flexibility in model selection. Regardless of the initial algorithm choice, PDx mitigates value erosion caused by model degradation through continuous retraining and adaptation. To establish this, we subjected our AI-driven end-to-end PD modeling pipeline to four well established machine learning algorithms: Logistic Regression (LR), Random Forest (RF), Extreme Gradient Boosting (XGBoost), and Neural Networks (NN). Detailed algorithmic descriptions and parameter considerations are provided in Appendix A. Given the critical role of model interpretability in digital lending, we prioritized variable selection to streamline the set of covariates used in model fitting. Furthermore, we utilized Optuna, a Bayesian optimization library, to fine-tune hyperparameters efficiently. Two key innovations within PDx—the integration of MLOps and the implementation of a champion-challenger framework—ensure automated model governance, adaptation, and sustained predictive accuracy, as detailed below.

\subsection{MLOps}

Initially inspired as a set of best engineering practices from DevOps, MLOps has grown into a holistic approach for managing the full Machine Learning Model lifecycle. This includes integrating with model development, orchestration, deployment, and monitoring model health, diagnostics, and governance, all while aligning with business goals. Although MLOps has been significantly applied in fields like image classification \cite{sitcheu2023mlops} and manufacturing \citep{testi2022mlops, raffin2022reference, lim2019mlop}, its use in finance, especially in consumer risk prediction, is still largely unexplored. This paper investigates the effective application of MLOps in deploying, managing, and scaling ML models for consumer credit risk management in digital lending, aiming to transform customer risk assessment methods. Digital lending involves customer data that can change due to various economic factors, leading to a decline in data drifts and model performance. MLOps combines Machine Learning with DevOps and data engineering practices to boost the efficiency of deploying and managing ML systems. It broadens the scope of traditional DevOps by including data and models, covering Continuous Integration (CI) for automatic building, testing, and validating source code; Continuous Training (CT) for automatic retraining and performance comparison of ML models; and Continuous Deployment (CD) for frequent automated model releases in production. In this paper, we established that an integrated approach is key to keeping pace with the fast-changing ML landscape and is poised to significantly enhance PD model development, deployment, and management, ultimately improving overall business performance. The following section explores the development of an algorithm (PDx) using MLOps practices to iteratively enhance ML models. These models are designed to integrate the latest data trends, challenge current baseline models, and emerge as new champions. By doing so, the PDx algorithm aims to safeguard against value erosion and significantly enhance the effectiveness of PD model predictions. \\

\subsection{PDx - Boosted PD model}
\begin{figure}
    \centering
    \includegraphics[width=0.9\linewidth]{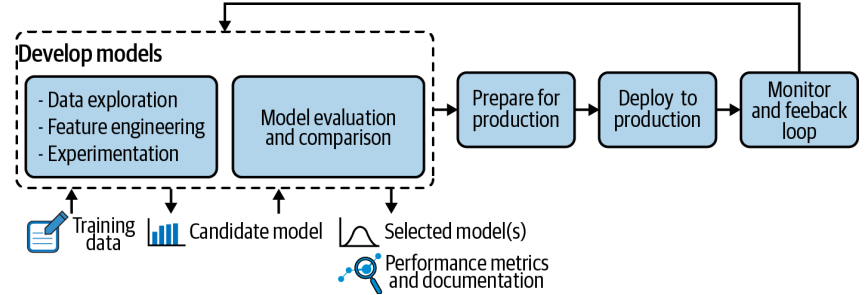}
    \caption{Model development highlighted in the larger context of the ML project life
cycle \cite{treveil2020introducing}}
    \label{fig:enter-label-1}
\end{figure}

The ML model lifecycle encompasses several crucial stages: data collection, processing, cleaning, and feature engineering. Subsequently, it progresses to model development, evaluation through competitive validation and selection, deployment, and ongoing monitoring, \citep{baesens2023boosting,treveil2020introducing,treveil2020introducing} as illustrated in Figure \ref{fig:enter-label-1}. However, a notable drawback of the current model development approach is the oversight of crucial aspects such as version control and proactive failure management. The current method for handling failures is reactive: model retraining occurs only after a significant performance decline. This reactive approach can lead to inefficiencies and delays in addressing performance issues, underscoring the necessity for a proactive and streamlined model development and management approach. To address these challenges, the PDx algorithm, using the best MLOps practices, proactively develops and continuously updates the probability of default model in production. This approach aims to improve performance and effectively combat value erosion.

    \begin{algorithm}
    \small
    \caption{Champion-challenger experiment design for production}\label{alg:cap}
    \begin{algorithmic}
    \State Champion model $M^\star \gets $ None
    \State Champion score $S^\star \gets 0$
    \For{each production month}
        \If{$M^\star \not= $ None} 
            \State $S^\star \gets$ Evaluate $M^\star$ on out-of-time-validation set $D_{OTV}^{t\left(i\right)}$
        \EndIf
        \For{each challenger model $M^\dagger$}
            \For{each time-series forecasting method $t\left(i\right)$}
                \State Train $M^\dagger$ on training set $D_{Train}^{t\left(i\right)}$, and tune the hyperparameters of $M^\dagger$ on in-time-validation set $D_{ITV}^{t\left(i\right)}$
                \State Challenger score $S^\dagger \gets$ Evaluate $M^\dagger$ on out-of-time-validation set $D_{OTV}^{t\left(i\right)}$
                \If{$S^\dagger > S^\star$} 
                    \State $S^\star \gets S^\dagger$
                    \State $M^\star \gets M^\dagger$
                \EndIf
            \EndFor
        \EndFor
    \EndFor
    \end{algorithmic}
    \end{algorithm}
    
\begin{figure*}
    \centering
    \includegraphics[width=\linewidth]{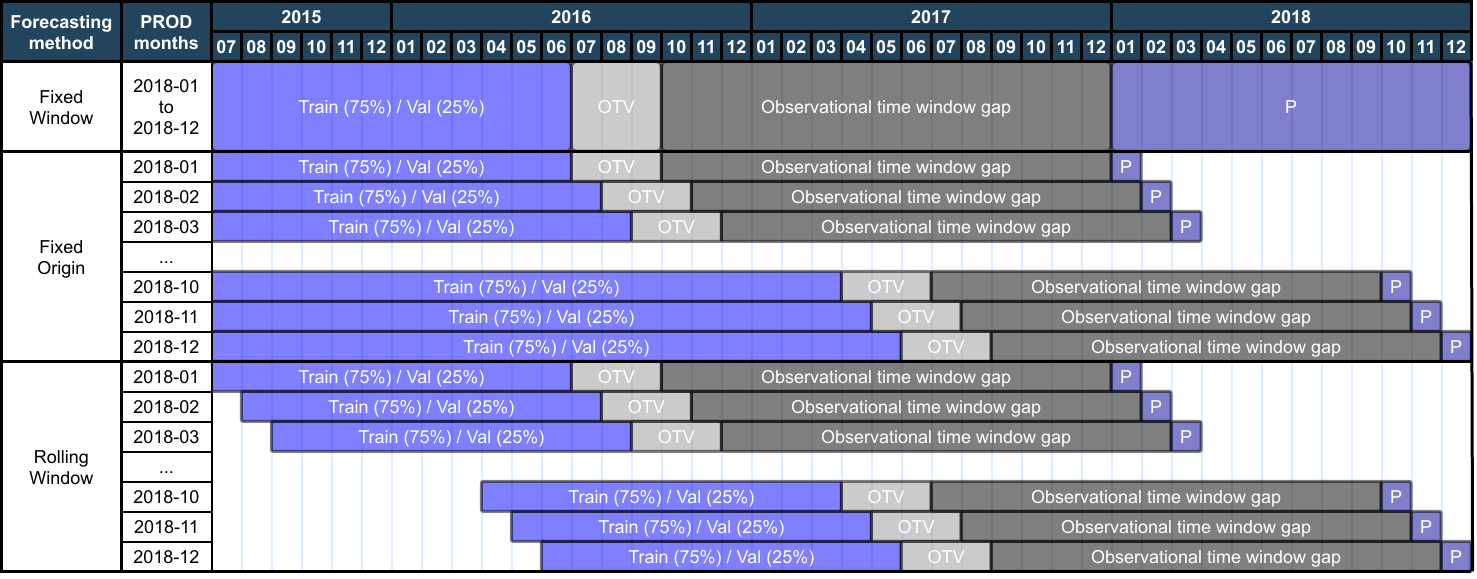}
    \caption{Conceptual approach for PDx experiment design (continuous training and validation) }
    \label{fig:forecasting-method-timesplits}
\end{figure*}

The PDx algorithm presented in (Algorithm \ref{alg:cap}) begins by identifying the initial champion model (best initial baseline model), denoted as \( M^\star \), through a comparative analysis of \textbf{fixed window model} performance against selected base algorithms such as Logistic Regression (LR), Random Forest (RF), XGBoost (XGB), and Neural Networks (NN). This champion model is then deployed in production during the first month and serves as a benchmark model in subsequent iterations to help estimate uplift from our proposed approach. Each time before deploying the model in the next production month, challenger models (example: Table \ref{tab:dev-table}) are constructed using the newly available training data (see Figure \ref{fig:forecasting-method-timesplits}). Also, the reigning champion model from the previous month undergoes a comparative evaluation against both newly developed and existing challenger models, denoted as \( M^\dagger \), using common testing data from the most recent available out-of-time validation window. The evaluation encompasses classification metrics such as AUC, 
and aspects of model interpretability such as feature importance and SHAP values, with human oversight integrated into the process. This rigorous assessment ensures the selection of the most effective model for deployment in the production environment.

Model updates are strategized by altering the learning algorithm or modifying the data input. To generate new challenger models at each stage, we employ the algorithms used in baseline model development and retrained them through two specific approaches:

\paragraph{\textbf{Fixed Origin Re-calibration}}
This method maintains a constant starting point for the development dataset, systematically incorporating new data as it becomes available. By retraining the PD models with each data addition, this strategy aims to enhance or maintain model performance by integrating the latest data trends.

\paragraph{\textbf{Rolling Window}}
The Rolling Window strategy dynamically adjusts the start point of the dataset, moving forward by the same width as the newly added data relative to the original development set. Concurrently, it incorporates fresh data while phasing out older, potentially outdated information. This method strives to refine model performance by leveraging recent data trends and discarding obsolete patterns that may no longer hold relevance. 

The application of PDx within the champion-challenger framework is elaborated in Figure \ref{fig:forecasting-method-timesplits}, showcasing its implementation on the P2P digital lending dataset as a crucial component of our experimental design.


\section{Experimental Design}
\label{sec:expt-design}
This section details the experiments conducted for our proposed PDx- adaptive credit risk forecasting model. It includes the dataset description, objective function, and target variable definition, criteria for model performance evaluation, training and validation methods, and the conceptual approach for deployment and monitoring used in this research.

\subsection{Data sets}
As credit scoring models deal with consumer data, data availability poses a challenge for credit scoring research due to privacy, business, and regulatory risks. Most studies rely on public datasets from the University of California (UCI) Machine Learning Repository \cite{boughaci2018new} or Kaggle community \citep{ariza2020explainability,he2023privacy} and only some on private data \citep{papouskova2019two,shen2020three} through industry-academics collaboration. These datasets typically comprise various features, including loan specifics, borrower's creditworthiness, payment history, credit inquiries, customer demographics, application information, alternative data sources, and a binary outcome for loan default. However, limitations such as small sample sizes often limit public datasets, as seen in UCI’s Japanese, German, and Australian datasets, which do not exceed 1,000 cases. Moreover, they typically lack time-related data, such as loan disbursal dates, which restricts the exploration of a model’s longevity and its performance over time, such as famous datasets like GMSC \citep{wang2024novel,zhang2019novel,anis2017investigating} and Home Credit \citep{wang2024novel, cocskun19credit} datasets from kaggle. Though rich in information, the private datasets used in previous research are bound by confidentiality agreements, limiting their reusability for future exploratory studies.

To evaluate our model effectively in a production setting, we utilized credit approval data spanning different periods rather than relying on a static snapshot. This approach enabled us to simulate realistic scenarios for training, testing, and validating the model on historical data and deploying it for future performance assessment in a live environment. Such a methodology allowed us to compare a baseline fixed model and our proposed PDx model. For a thorough validation of PDx algorithms' performance, this study incorporated three diverse credit datasets summarised in table \ref{tab:datasets-metadata}, varying in size, loan types, and target variables, ensuring a comprehensive analysis and robust results.

\begin{itemize}
    \item \textbf{P2P Dataset}: The first dataset the study utilizes is from Lending Club, one of the leading peer-to-peer (P2P) digital personal loan providers in the U.S. We have used one-year post-disbursal repayment pattern to identify customers who would reach 90 days past due (DPD) within the first 12 months of loan start date. This dataset is regularly updated and accessible on the Lending Club website for research purposes. Numerous studies \citep{xia2017cost, papouskova2019two, zhang2020credit, zhu2023explainable, ariza2020explainability, croux2020important,serrano2016use, guo2016instance, serrano2015determinants} have analyzed this dataset to pinpoint the optimal analytical methods for credit scoring, identify best classifiers, key variables, and profit maximization strategies. However, many of these investigations treated the dataset as static, overlooking the impact of temporal data shifts, operational failures, and the need for model upkeep as potential solutions. To address these gaps and strike a balance between data recency and completeness, we used the data with loan start dates from July'15 until Dec'18, and since the target variable used on this dataset is 90 DPD in 12 months' last payment date for loans disbursed in dec'18 is Dec'19. The dataset comes with 150 features for each loan, including attributes from loan applications (e.g., date of application, type of loan, the reason for the loan, applied amount, tenure, etc. ), credit bureaus (e.g., past and active tradelines, loan inquiry information, credit score, and grades), customer demographic (address,  age, income, home ownership, employment, debt to income ratio, etc.), and repayment information (e.g., payment start date, hardship, last payment date, settlement, loan status, etc.)  
    
    \item \textbf{SBA Dataset}: The second dataset that this study leverages is a comprehensive dataset from the U.S. Small Business Administration (SBA), as detailed in previous research work \citep{li2018should, glennon2005measuring}. The dataset contains over 900K disbursed loan data since 1961. For this study, we used loans approved and disbursed between 1998 and 2004. 75\% of data from 1998 to 2000 serves as our development or training set, with 25\% of these loans allocated for in-time validation. The first quarter of 2001 is used for out-of-time (OTV) validation to assist in baseline model selection. Data from the first quarter of 2002 through to the end of 2004 represents the model's production period in our study. The dataset has 26 variables for each loan, including attributes from loan applications, demographics, business details (e.g., number of employees, business type, etc.), and bureau data.

    \item \textbf{AutoL Dataset}: The third dataset used for this research is private data from a two-wheeler vehicle loan (auto loan)  provider digital lender in India and procured from a machine learning competition hosted by Analytics Vidhya. The target variable used is 30 DPD in 3 months from the loan start date. The dataset includes information from application data, demographics,  bureau data, asset details, and post-disbursal payments.  
    
\end{itemize}

The datasets are partitioned into segments to mimic a real credit scoring application scenario, including model training, testing, validation, and productionization. Several preprocessing steps were implemented to prepare the dataset. First, dates were transformed into the number of days from a reference date (e.g., date of application), such as converting \texttt{last\_pymnt\_d} to \texttt{num\_days\_payment} by calculating the difference between \texttt{last\_pymnt\_d} and \texttt{issue\_d} or application date. Next, columns prone to data leakage or future information, like \texttt{pymnt\_plan}, \texttt{funded\_amnt}, \texttt{payment\_differment}, and \texttt{int\_rate}, were removed to maintain model integrity. Missing values were filled with zeros and flagged in a separate column to preserve data quality. Categorical variables were one-hot encoded, and infrequent categories (occurring less than 1\% of the time) were grouped into an ‘Others’ category. Standardization was applied solely to the training set to prevent data leakage, and this scaling method was consistently applied across all data splits (Train, ITV, OTV, PROD). Finally, all rows were retained post-inspection for anomalies and outliers.

\begin{table}[htbp]
\centering
\setlength\tabcolsep{3pt}
\caption[]{Metadata of datasets used}
\label{tab:datasets-metadata}
{
\begin{tabular}{cccccc}
\hline
\multirow{2}{*}{Datasets} & \multirow{2}{*}{No. of features} & \multicolumn{4}{c}{\begin{tabular}[c]{@{}c@{}}No. of observations \\ (bad rate)\end{tabular}} \\
                          &                                  & Train                 & ITV                   & OTV                   & PROD                  \\ \hline
P2P &
  150 &
  \begin{tabular}[c]{@{}c@{}}354,551 \\ (18.6\%)\end{tabular} &
  \begin{tabular}[c]{@{}c@{}}118,183 \\ (18.6\%)\end{tabular} &
  \begin{tabular}[c]{@{}c@{}}99,120 \\ (18.9\%)\end{tabular} &
  \begin{tabular}[c]{@{}c@{}}36,347 \\ (11.8\%)\end{tabular} \\
\multicolumn{1}{l}{}      & \multicolumn{1}{l}{}             & \multicolumn{1}{l}{}  & \multicolumn{1}{l}{}  & \multicolumn{1}{l}{}  & \multicolumn{1}{l}{}  \\
SBA &
  26 &
  \begin{tabular}[c]{@{}c@{}}44,030 \\ (5.5\%)\end{tabular} &
  \begin{tabular}[c]{@{}c@{}}14,677 \\ (5.5\%)\end{tabular} &
  \begin{tabular}[c]{@{}c@{}}4,592 \\ (6.0\%)\end{tabular} &
  \begin{tabular}[c]{@{}c@{}}5,109 \\ (5.7\%)\end{tabular} \\
\multicolumn{1}{l}{}      & \multicolumn{1}{l}{}             & \multicolumn{1}{l}{}  & \multicolumn{1}{l}{}  & \multicolumn{1}{l}{}  & \multicolumn{1}{l}{}  \\
AutoL &
  46 &
  \begin{tabular}[c]{@{}c@{}}37,954 \\ (17.7\%)\end{tabular} &
  \begin{tabular}[c]{@{}c@{}}12,651 \\ (17.7\%)\end{tabular} &
  \begin{tabular}[c]{@{}c@{}}12,573 \\ (17.3\%)\end{tabular} &
  \begin{tabular}[c]{@{}c@{}}4,188 \\ (17.1\%)\end{tabular} \\ \hline
\end{tabular}
}
\end{table}

\subsection{Performance evaluation metrics}
To assess the predictive power and model stability
we employed standard classification metrics as referenced in Table \ref{table:previous-studies}. 
Our evaluation of predictive power utilized metrics such as the Area Under the Curve (AUC), Kolmogorov-Smirnov (KS),  F1-Score, and Recall as 3 decile capture rate. The foundation for evaluating a classification model's predictive power lies in the confusion matrix: 
\[
\begin{array}{cc}
 & \text{Predicted} \\
\text{Actual} &
\begin{bmatrix}
TN & FP \\
FN & TP
\end{bmatrix}
\end{array}
\]

Where \( TP \), \( TN \), \( FP \), and \( FN \) denote True Positives, True Negatives, False Positives, and False Negatives, respectively. For each instance in the dataset, the model assigns a score, which, when sorted, serves as a cutoff or threshold, leading to a revised confusion matrix. The \textbf{AUC} metric is calculated as the integral of the True Positive Rate (TPR) over the False Positive Rate (FPR) across varying threshold values:

\[
\text{AUC} = \int_{0}^{1} \text{TPR}(FPR) \, d(FPR)
\]

where: 
    \( \text{TPR}(FPR) \) represents the True Positive Rate as a function of the False Positive Rate with \( \text{TPR} = \frac{TP}{TP + FN} \)
    and \( \text{FPR} = \frac{FP}{FP + TN} \) . 

\textbf{F1-score} brings the balanced view of precision (P) and recall (R) and is defined as the harmonic average between them:  \[ F1 = 2 \times \frac{P \times R}{P + R} \] where \( \text{P} = \frac{TP}{TP + FP} \) and \( \text{R} = \frac{TP}{TP + FN} \). \textbf{KS} metric, representing the maximal disparity between the cumulative distributions of events and non-events, is defined as \[ \text{KS} = \max_{t} | \text{TPR}(t) - \text{FPR}(t) | \] where \( \max_{t} \) represents the maximum difference across all thresholds \( t \). The theoretical values of these metrics range from 0 to 1, where 1 indicates the best possible performance and the model's ability to separate the good and bad loan applications perfectly. We have used the top 3 decile capture (bad rate) to access model stability.

\subsection{Model training and validation}

To thoroughly highlight the superiority of the proposed learning framework, we adhered to best practices in model training and validation. The process began with inputting data that underwent rigorous cleaning and preprocessing, including optimal techniques for missing value imputation and variable standardization. Variable selection was refined through iterative processes, complemented by hyperparameter optimization via Bayesian methods \cite{turner2021bayesian}. As we have used multiple algorithms, models from each iteration were compared and evaluated based on AUC, PSI, and Equal Opportunity metrics on validation datasets to select the best base model. Further, during the champion/challenger environment with MLOPs, for continuous model improvement, a similar methodology was applied to train and validate models to newly available training and validation data, ensuring the identification and deployment of updated champion models for production. For the P2P dataset, this approach has been explained in fig \ref{fig:forecasting-method-timesplits}.


\begin{figure}
    \centering
    \includegraphics[scale = 0.5]{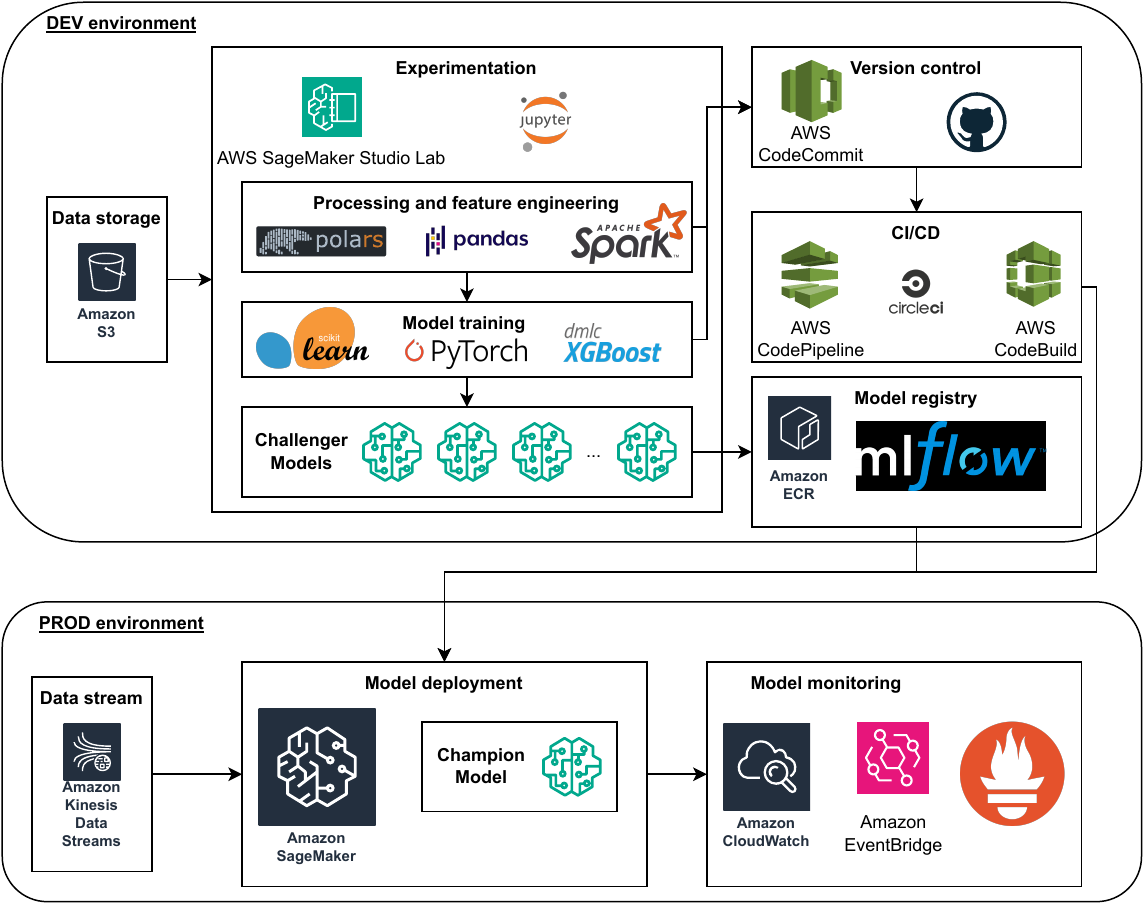}
    \caption{Conceptual  architecture for PDx deployment on AWS.}
    \label{fig:enter-label}
\end{figure}

\subsection{Conceptual deployment and monitoring architecture}

We simulated the experiments as an actual deployment setting, mainly using services provided by AWS Cloud as seen in fig \ref{fig:enter-label}. The codes are implemented in Python version 3.11 in both the development and production environments of Amazon Sagemaker. Data stored in Amazon S3 is structured to adhere to the layered data-engineering convention.

Each dataset is divided into three parts - development set, OTV set, and production set. The development set comprises the training set (75\%) and in-time validation set (25\%), which are used for training and fine-tuning the models. The OTV set is used to select the champion model for deployment in the production environment. AWS Batch is used to scheduling a monthly/quarterly batch processing job run that pulls the development set and OTV set from the Amazon S3 data lake and feeds it to the training pipelines. To ensure reproducibility, these model training runs are tracked in MLflow together with its version, parameters, metrics, and artifacts. The champion model is packaged into a Docker container image to be stored in Amazon ECR.

A time gap is enforced between the OTV set and production set since customers in the development set and OTV set would require an observational window period to be labeled according to the default definition. Data from the production set is continuously fed to the champion model SageMaker endpoint using Amazon Kinesis Data Streams. To ensure reliable real-time operations, we leveraged Amazon CloudWatch Anomaly Detection for drift detection, Amazon SNS for automated notifications, and AWS Lambda for triggering the model pipelines for retraining.

\section{Results and Data Analysis}
\label{sec:results}
This section details the findings from the numerous experiments carried out during the research.


\begin{table}[ht]
\centering
\caption{Baseline model results in development}
\label{tab:dev-table-all}
{
\begin{tabular}{cclllll}
\hline
\multirow{2}{*}{\textbf{\begin{tabular}[c]{@{}c@{}}Data \\ Sets\end{tabular}}} &
  \multirow{2}{*}{\textbf{\begin{tabular}[c]{@{}c@{}}Used \\ Algorithm\end{tabular}}} &
  \multicolumn{4}{c}{\textbf{Performance Metric (ITV, OTV)}} \\
 &
 &
  \multicolumn{1}{c}{\textbf{AUC}} &
  \multicolumn{1}{c}{\textbf{KS}} &
  \multicolumn{1}{c}{\textbf{F-1}} &
  \multicolumn{1}{c}{\textbf{3DCR}} \\ \hline
                & LR          & 0.702, 0.682 & 0.290, 0.256 & 0.402, 0.354 & 0.526, 0.499 \\
                & RF          & 0.701, 0.678 & 0.288, 0.241 & 0.401, 0.347 & 0.523, 0.488 \\
P2P             & XGB         & {\color[HTML]{0000FF} 0.705, 0.687} & {\color[HTML]{0000FF} 0.292, 0.264} & {\color[HTML]{0000FF} 0.403, 0.361} & {\color[HTML]{0000FF} 0.527, 0.508}  \\
                & NN          & 0.694, 0.655 & 0.280, 0.214 & 0.394, 0.323 & 0.517, 0.455 \\ \hline

                & LR          & 0.757, 0.748 & 0.372, 0.377 & 0.203, 0.212 & 0.651, 0.655 \\
                & RF          & 0.879, 0.882 & 0.611, 0.585 & 0.274, 0.276 & 0.877, 0.851 \\
SBA             & XGB         & {\color[HTML]{0000FF} 0.924, 0.936} & {\color[HTML]{0000FF} 0.716, 0.684} & {\color[HTML]{0000FF} 0.285, 0.293} & {\color[HTML]{0000FF} 0.913, 0.902}  \\
                & NN          & 0.796, 0.785 & 0.487, 0.389 & 0.232, 0.216 & 0.743, 0.665 \\ \hline

                & LR          & 0.700, 0.664 & 0.306, 0.170 & 0.409, 0.323 & 0.548, 0.434 \\
                & RF          & {\color[HTML]{0000FF} 0.725, 0.687} & {\color[HTML]{0000FF} 0.331, 0.216} & {\color[HTML]{0000FF} 0.418, 0.348} & {\color[HTML]{0000FF} 0.560, 0.468} \\
AutoL           & XGB         & 0.684, 0.650 & 0.279, 0.184 & 0.384, 0.331 & 0.515, 0.445 \\
                & NN          & 0.731, 0.681 & 0.344, 0.199 & 0.435, 0.339 & 0.582, 0.455 \\ \hline

\end{tabular}
}
\end{table}

We begin our experiments by identifying the optimal baseline model for predicting customer probability of default (PD) scores on future loan applications. This serves as the initial champion model selection, achieved through model training on available datasets using well-established classifiers, including Logistic Regression (LR), Extreme Gradient Boosting (XGBoost), Random Forest (RF), and Neural Networks (NN). This step also allowed us to identify key limitation in existing research, where models are typically deployed to production after initial selection and remain unchanged for nearly a year before the next update cycle. 

Across the three datasets, XGBoost consistently outperformed LR and NN in the validation data, consistent with the findings of previous studies \cite{gunnarsson2021deep}. However, RF demonstrated better performance on the AutoL data set in terms of AUC when evaluated on the future out of time validation data (OTV), as shown in Table \ref{tab:dev-table-all}. This highlights our first key insight: while certain algorithms serve as strong baselines, the optimal model is ultimately a function of both data characteristics and algorithm selection. Therefore, systematically exploring a diverse set of well-established algorithms can provide incremental gains in predictive performance, ensuring the most effective model is chosen for deployment. 

\begin{figure}
      \centering
      \includegraphics[width=0.75\linewidth]{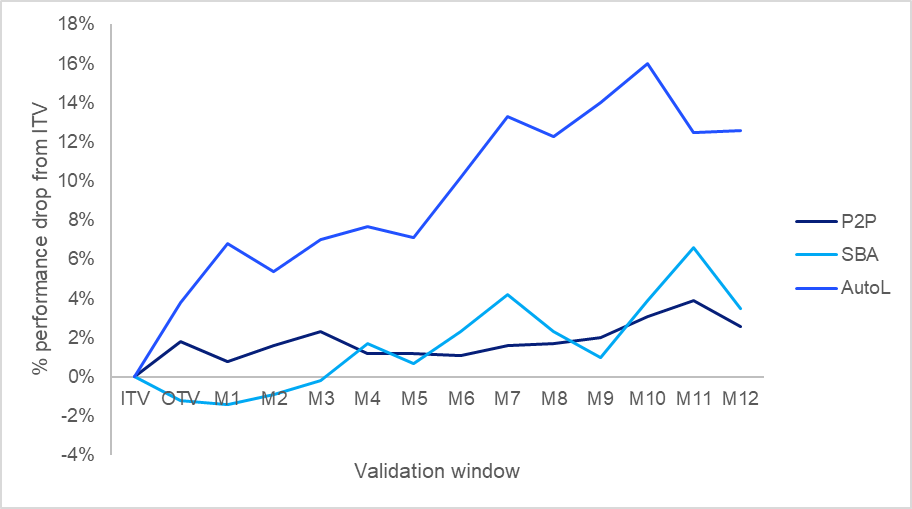}
      \caption{Model performance drop in production }
      \label{fig:model-performancedrop}
  \end{figure}

To ensure an unbiased comparison, we standardized ML pipeline development by adhering to best practices in feature standardization, feature engineering, feature reduction, variable selection, and hyperparameter tuning during each model training process. Additionally, all models were built with interpretability and reproducibility in mind. Our second key finding: all models, regardless of the baseline algorithm, experience performance degradation if left unupdated. This is evident in Tables \ref{tab:prod-table1}, \ref{tab:prod-table2}, and \ref{tab:prod-table3}, where the fixed-window results across the datasets demonstrate a significant performance decline over time. This is applicable even if we select best model after assessing multiple algorithms. As we observed, though XGBoost (XGB) found to be the most effective model for both the P2P and SBA scenarios and Random Forest (RF) as the champion in the AutoLoan experiment, as detailed in Table \ref{tab:dev-table-all}, however, as depicted in Figure \ref{fig:model-performancedrop}, a significant drop in model performance was noted following deployment, with AUC reductions of 4\%, 7\%, and 16\% observed in the P2P, SBA, and AutoL experiments, as well. These findings highlight the critical need for continuous monitoring of both model performance and loan disbursement outcomes to accurately compare predicted versus actual default rates over the designated observation period. Without timely interventions, performance degradation may worsen, exacerbating existing challenges and potentially leading to an increase in non-performing assets (NPAs) across lending portfolios. This aspect of post-deployment model deterioration remains underexplored in existing research, emphasizing the need for further investigation into strategies for sustained model adaptability and risk mitigation. 

PDx delivers promising results in addressing these challenges by implementing a dynamic, competition-driven modeling strategy. In our framework PDx periodically generates challenger models using \textbf{fixed-origin recalibration} and \textbf{rolling-window} data updates, ensuring that the champion model remains competitive and continuously optimized in production. This iterative process of data refreshing, model retraining, comparing, and selection enhances overall prediction efficacy by systematically managing competition among models. PDx demonstrated consistent good results in maintaining predictive power across our P2P, SBA, and AutoL experiments over time. In the P2P digital lending experiment, the continuous development of challenger models, and best champion model selection as presented in Table \ref{tab:dev-table}, highlight the effectiveness of continuous learning in a competitive setup. Using the most current data, PDx maintains predictive accuracy and adapts to evolving borrower behavior. Our result also shows that after the first deployment of model in production, the original fixed window model ceased to dominate as the champion, replaced by either fixed origin recalibrated or rolling window updated models in subsequent iterations, as evidenced by the champion models listed in Tables \ref{tab:prod-table1}, \ref{tab:prod-table2}, and \ref{tab:prod-table3} within the PDx production framework. Furthermore, in the Auto Loan experiment (Table \ref{tab:prod-table3}), the initial Random Forest champion was later outperformed by XGBoost, demonstrating that the best-performing algorithm can shift as new data becomes available, suggesting algo selection should not be static. Additionally, both fixed-origin recalibration and rolling-window approaches produced champion models across all experiments, highlighting the need for dataset-specific selection of data update mechanisms. These findings emphasize the importance of ongoing model evaluation and adaptive learning to sustain predictive performance in dynamic lending environments.

The tables referenced above, and (Table \ref{tab:dev-table-all-2}) clearly demonstrate the effectiveness of the PDx algorithm in production results by stopping performance degradation and and impact by capturing defaulters more accurately by 12th production iterations. In (Table \ref{tab:dev-table-all-2}), we can also see in the realm of P2P digital lending, the algorithm facilitated a 2\% enhancement in model efficacy leading to similar level of improvement in business decisions by capturing additional 2.2\% defaulters in 12th month's iteration from first launch. Similarly, in the SBA experiment, PDx achieved an AUC improvement exceeding 5\%, leading to a 9.8\% increase in default capture within the top three deciles of the risk distribution. In the Auto Loan experiment, model performance improved by 3.1\%, enabling a 9.2\% increase in capturing bad loans. These incremental performance gains across multiple metrics in the 12th production cycle highlight the long-term advantages of continuous model updates. We also observed similar performance improvement in other key metrics used in banking analytics like KS Score, F-1 score, etc.  Unlike conventional fixed-window approaches, which dominate both academic and industry practices, PDx systematically enhances predictive power through adaptive learning, reinforcing its superiority over static model selection methodologies. This significant reduction in risky approvals has strong potential to improve portfolio's overall quality, leading to a decrease in nonperforming asset (NPA) expenses and a positive impact on the PnL.

\begin{table}[ht]
\centering
\caption{PDx results in production - 12th iteration}
\label{tab:dev-table-all-2}
{
\begin{tabular}{cclllll}
\hline
\multirow{2}{*}{\textbf{\begin{tabular}[c]{@{}c@{}}Data \\ Sets\end{tabular}}} &
  \multirow{2}{*}{\textbf{\begin{tabular}[c]{@{}c@{}} Champion \\ Algorithm\end{tabular}}} &
  \multicolumn{5}{c}{\textbf{Model performance in Production}} \\
 &
 &
  \multicolumn{1}{c}{\textbf{AUC}} &
  \multicolumn{1}{c}{\textbf{KS}} &
  \multicolumn{1}{c}{\textbf{F-1}} &
  \multicolumn{1}{c}{\textbf{3DCR}} \\ \hline
                & FW XGB          & 0.679 & 0.272 & 0.256 & 0.532 \\
P2P (M12)             & PDx             & {\color[HTML]{0000FF} 0.701} & {\color[HTML]{0000FF} 0.295} & {\color[HTML]{0000FF} 0.267} & {\color[HTML]{0000FF} 0.554}  \\ 
                & ($\Delta$)      & 2.2\% & 2.3\% & 1.1\% & 2.2\% \\\hline

                & FW XGB          & 0.889 & 0.611 & 0.295 & 0.874 \\
SBA (Q12)            & PDx         & {\color[HTML]{0000FF} 0.948} & {\color[HTML]{0000FF} 0.709} & {\color[HTML]{0000FF} 0.326} & {\color[HTML]{0000FF} 0.966}  \\
                & ($\Delta$)          & 5.9\% & 9.8\% & 0.402 & 9.8\% \\ \hline

                & FW RF           & 0.599 & 0.157 & 0.312 & 0.422 \\
AutoL(M12)           & PDx          & {\color[HTML]{0000FF} 0.725} & {\color[HTML]{0000FF} 0.241} & {\color[HTML]{0000FF} 0.365} & {\color[HTML]{0000FF} 0.494} \\ 
                & ($\Delta$)          & 3.1\% & 9.8\% & 3.1\% & 9.2\% \\ \hline

\end{tabular}
}
\end{table}

\begin{table}[htbp]
\centering
\setlength\tabcolsep{3pt}
\caption[]{AUC of challenger models in development. Selected challenger models are highlighted in bold, deep blue.}
\label{tab:dev-table}
\resizebox{0.48\textwidth}{!} 
{
\begin{tabular}{ccccclccc}
\hline
 &
   &
  \multicolumn{3}{c}{\textbf{Fixed origin re-calibration}} &
   &
  \multicolumn{3}{c}{\textbf{Rolling window}} \\
\multirow{-2}{*}{\textbf{\begin{tabular}[c]{@{}c@{}}Production\\ (month - 1) \end{tabular}}} &
  \multirow{-2}{*}{\textbf{\begin{tabular}[c]{@{}c@{}}Challenger \\ models\end{tabular}}} &
  Train &
  ITV &
  {\color[HTML]{000000} OTV} &
   &
  Train &
  ITV &
  {\color[HTML]{000000} OTV} \\ \hline
0 &
  LR &
  0.701 &
  0.702 &
  {\color[HTML]{000000} 0.682} &
   &
  0.701 &
  0.702 &
  {\color[HTML]{000000} 0.682} \\
 &
  RF &
  0.769 &
  0.701 &
  {\color[HTML]{000000} 0.678} &
   &
  0.769 &
  0.701 &
  {\color[HTML]{000000} 0.678} \\
 &
  XGB &
  0.759 &
  0.705 &
  {\color[HTML]{0000FF} \textbf{0.687}} &
   &
  0.759 &
  0.704 &
  {\color[HTML]{0000FF} \textbf{0.687}} \\
 &
  NN &
  0.816 &
  0.694 &
  {\color[HTML]{000000} 0.655} &
   &
  0.816 &
  0.694 &
  {\color[HTML]{000000} 0.655} \\
\multicolumn{1}{l}{} &
  \multicolumn{1}{l}{} &
  \multicolumn{1}{l}{} &
  \multicolumn{1}{l}{} &
  \multicolumn{1}{l}{} &
   &
  \multicolumn{1}{l}{} &
  \multicolumn{1}{l}{} &
  \multicolumn{1}{l}{} \\
1 &
  LR &
  0.700 &
  0.699 &
  {\color[HTML]{000000} 0.682} &
   &
  0.700 &
  0.698 &
  {\color[HTML]{000000} 0.682} \\
 &
  RF &
  0.779 &
  0.699 &
  {\color[HTML]{000000} 0.679} &
   &
  0.775 &
  0.696 &
  {\color[HTML]{000000} 0.678} \\
 &
  XGB &
  0.757 &
  0.703 &
  {\color[HTML]{0000FF} \textbf{0.687}} &
   &
  0.760 &
  0.702 &
  {\color[HTML]{0000FF} 0.686} \\
 &
  NN &
  0.802 &
  0.698 &
  {\color[HTML]{000000} 0.621} &
   &
  0.812 &
  0.696 &
  {\color[HTML]{000000} 0.612} \\
\multicolumn{1}{l}{} &
  \multicolumn{1}{l}{} &
  \multicolumn{1}{l}{} &
  \multicolumn{1}{l}{} &
  \multicolumn{1}{l}{} &
   &
  \multicolumn{1}{l}{} &
  \multicolumn{1}{l}{} &
  \multicolumn{1}{l}{} \\
2 &
  LR &
  0.699 &
  0.700 &
  {\color[HTML]{000000} 0.680} &
   &
  0.698 &
  0.695 &
  {\color[HTML]{000000} 0.681} \\
 &
  RF &
  0.775 &
  0.699 &
  {\color[HTML]{000000} 0.678} &
   &
  0.780 &
  0.694 &
  {\color[HTML]{000000} 0.678} \\
 &
  XGB &
  0.753 &
  0.705 &
  {\color[HTML]{0000FF} 0.687} &
   &
  0.712 &
  0.704 &
  {\color[HTML]{0000FF} \textbf{0.690}} \\
 &
  NN &
  0.804 &
  0.697 &
  {\color[HTML]{000000} 0.623} &
   &
  0.821 &
  0.686 &
  {\color[HTML]{000000} 0.624} \\
\multicolumn{1}{l}{} &
  \multicolumn{1}{l}{} &
  \multicolumn{1}{l}{} &
  \multicolumn{1}{l}{} &
  \multicolumn{1}{l}{} &
   &
  \multicolumn{1}{l}{} &
  \multicolumn{1}{l}{} &
  \multicolumn{1}{l}{} \\
3 &
  LR &
  0.698 &
  0.700 &
  {\color[HTML]{000000} 0.676} &
   &
  0.698 &
  0.694 &
  {\color[HTML]{000000} 0.677} \\
 &
  RF &
  0.774 &
  0.698 &
  {\color[HTML]{000000} 0.674} &
   &
  0.783 &
  0.695 &
  {\color[HTML]{000000} 0.679} \\
 &
  XGB &
  0.752 &
  0.706 &
  {\color[HTML]{0000FF} \textbf{0.687}} &
   &
  0.758 &
  0.696 &
  {\color[HTML]{0000FF} 0.682} \\
 &
  NN &
  0.799 &
  0.701 &
  {\color[HTML]{000000} 0.647} &
   &
  0.826 &
  0.682 &
  {\color[HTML]{000000} 0.642} \\
\multicolumn{1}{l}{} &
  \multicolumn{1}{l}{} &
  \multicolumn{1}{l}{} &
  \multicolumn{1}{l}{} &
  \multicolumn{1}{l}{} &
   &
  \multicolumn{1}{l}{} &
  \multicolumn{1}{l}{} &
  \multicolumn{1}{l}{} \\
4 &
  LR &
  0.696 &
  0.701 &
  {\color[HTML]{000000} 0.678} &
   &
  0.695 &
  0.696 &
  {\color[HTML]{000000} 0.678} \\
 &
  RF &
  0.762 &
  0.701 &
  {\color[HTML]{000000} 0.675} &
   &
  0.773 &
  0.695 &
  {\color[HTML]{000000} 0.675} \\
 &
  XGB &
  0.750 &
  0.708 &
  {\color[HTML]{0000FF} \textbf{0.687}} &
   &
  0.763 &
  0.701 &
  {\color[HTML]{0000FF} 0.686} \\
 &
  NN &
  0.787 &
  0.704 &
  {\color[HTML]{000000} {0.638}} &
   &
  0.818 &
  0.688 &
  {\color[HTML]{000000} 0.630} \\
\multicolumn{1}{l}{} &
  \multicolumn{1}{l}{} &
  \multicolumn{1}{l}{} &
  \multicolumn{1}{l}{} &
  \multicolumn{1}{l}{} &
   &
  \multicolumn{1}{l}{} &
  \multicolumn{1}{l}{} &
  \multicolumn{1}{l}{} \\
5 &
  LR &
  0.696 &
  0.699 &
  {\color[HTML]{000000} 0.677} &
   &
  0.696 &
  0.690 &
  {\color[HTML]{000000} 0.678} \\
 &
  RF &
  0.768 &
  0.699 &
  {\color[HTML]{000000} 0.674} &
   &
  0.771 &
  0.690 &
  {\color[HTML]{000000} 0.674} \\
 &
  XGB &
  0.744 &
  0.704 &
  {\color[HTML]{0000FF} 0.685} &
   &
  0.709 &
  0.699 &
  {\color[HTML]{0000FF} \textbf{0.688}} \\
 &
  NN &
  0.789 &
  0.701 &
  {\color[HTML]{000000} 0.648} &
   &
  0.825 &
  0.680 &
  {\color[HTML]{000000} 0.636} \\
\multicolumn{1}{l}{} &
  \multicolumn{1}{l}{} &
  \multicolumn{1}{l}{} &
  \multicolumn{1}{l}{} &
  \multicolumn{1}{l}{} &
   &
  \multicolumn{1}{l}{} &
  \multicolumn{1}{l}{} &
  \multicolumn{1}{l}{} \\
6 &
  LR &
  0.696 &
  0.694 &
  {\color[HTML]{000000} 0.680} &
   &
  0.691 &
  0.693 &
  {\color[HTML]{000000} 0.681} \\
 &
  RF &
  0.744 &
  0.695 &
  {\color[HTML]{000000} 0.679} &
   &
  0.755 &
  0.689 &
  {\color[HTML]{000000} 0.680} \\
 &
  XGB &
  0.753 &
  0.703 &
  {\color[HTML]{0000FF} \textbf{0.692}} &
   &
  0.756 &
  0.697 &
  {\color[HTML]{0000FF} 0.692} \\
 &
  NN &
  0.782 &
  0.702 &
  {\color[HTML]{000000} 0.619} &
   &
  0.823 &
  0.681 &
  {\color[HTML]{000000} 0.621} \\
\multicolumn{1}{l}{} &
  \multicolumn{1}{l}{} &
  \multicolumn{1}{l}{} &
  \multicolumn{1}{l}{} &
  \multicolumn{1}{l}{} &
   &
  \multicolumn{1}{l}{} &
  \multicolumn{1}{l}{} &
  \multicolumn{1}{l}{} \\
7 &
  LR &
  0.694 &
  0.696 &
  {\color[HTML]{000000} 0.681} &
   &
  0.690 &
  0.690 &
  {\color[HTML]{000000} 0.682} \\
 &
  RF &
  0.792 &
  0.701 &
  {\color[HTML]{000000} 0.685} &
   &
  0.774 &
  0.691 &
  {\color[HTML]{000000} 0.685} \\
 &
  XGB &
  0.757 &
  0.700 &
  {\color[HTML]{0000FF} 0.690} &
   &
  0.756 &
  0.692 &
  {\color[HTML]{0000FF} \textbf{0.692}} \\
 &
  NN &
  0.779 &
  0.704 &
  {\color[HTML]{000000} 0.642} &
   &
  0.825 &
  0.677 &
  {\color[HTML]{000000} 0.623} \\
\multicolumn{1}{l}{} &
  \multicolumn{1}{l}{} &
  \multicolumn{1}{l}{} &
  \multicolumn{1}{l}{} &
  \multicolumn{1}{l}{} &
   &
  \multicolumn{1}{l}{} &
  \multicolumn{1}{l}{} &
  \multicolumn{1}{l}{} \\
8 &
  LR &
  0.695 &
  0.693 &
  {\color[HTML]{000000} 0.678} &
   &
  0.689 &
  0.688 &
  {\color[HTML]{000000} 0.679} \\
 &
  RF &
  0.781 &
  0.696 &
  {\color[HTML]{000000} 0.681} &
   &
  0.763 &
  0.688 &
  {\color[HTML]{000000} 0.679} \\
 &
  XGB &
  0.732 &
  0.700 &
  {\color[HTML]{0000FF} \textbf{0.696}} &
   &
  0.750 &
  0.692 &
  {\color[HTML]{0000FF} 0.690} \\
 &
  NN &
  0.770 &
  0.704 &
  {\color[HTML]{000000} 0.638} &
   &
  0.819 &
  0.681 &
  {\color[HTML]{000000} 0.623} \\
\multicolumn{1}{l}{} &
  \multicolumn{1}{l}{} &
  \multicolumn{1}{l}{} &
  \multicolumn{1}{l}{} &
  \multicolumn{1}{l}{} &
   &
  \multicolumn{1}{l}{} &
  \multicolumn{1}{l}{} &
  \multicolumn{1}{l}{} \\
9 &
  LR &
  0.700 &
  0.700 &
  {\color[HTML]{000000} 0.692} &
   &
  0.689 &
  0.686 &
  {\color[HTML]{000000} 0.688} \\
 &
  RF &
  0.782 &
  0.700 &
  {\color[HTML]{000000} 0.683} &
   &
  0.775 &
  0.686 &
  {\color[HTML]{000000} 0.684} \\
 &
  XGB &
  0.730 &
  0.703 &
  {\color[HTML]{0000FF} \textbf{0.700}} &
   &
  0.762 &
  0.689 &
  {\color[HTML]{0000FF} 0.697} \\
 &
  NN &
  0.769 &
  0.707 &
  {\color[HTML]{000000} 0.661} &
   &
  0.820 &
  0.675 &
  {\color[HTML]{000000} 0.662} \\
\multicolumn{1}{l}{} &
  \multicolumn{1}{l}{} &
  \multicolumn{1}{l}{} &
  \multicolumn{1}{l}{} &
  \multicolumn{1}{l}{} &
   &
  \multicolumn{1}{l}{} &
  \multicolumn{1}{l}{} &
  \multicolumn{1}{l}{} \\
10 &
  LR &
  0.699 &
  0.702 &
  {\color[HTML]{000000} 0.693} &
   &
  0.687 &
  0.690 &
  {\color[HTML]{000000} 0.690} \\
 &
  RF &
  0.779 &
  0.700 &
  {\color[HTML]{000000} 0.683} &
   &
  0.777 &
  0.689 &
  {\color[HTML]{000000} 0.686} \\
 &
  XGB &
  0.730 &
  0.706 &
  {\color[HTML]{0000FF} 0.701} &
   &
  0.707 &
  0.702 &
  {\color[HTML]{0000FF} \textbf{0.704}} \\
 &
  NN &
  0.765 &
  0.709 &
  {\color[HTML]{000000} 0.682} &
   &
  0.834 &
  0.673 &
  {\color[HTML]{000000} 0.670} \\
\multicolumn{1}{l}{} &
  \multicolumn{1}{l}{} &
  \multicolumn{1}{l}{} &
  \multicolumn{1}{l}{} &
  \multicolumn{1}{l}{} &
   &
  \multicolumn{1}{l}{} &
  \multicolumn{1}{l}{} &
  \multicolumn{1}{l}{} \\
12 &
  LR &
  0.699 &
  0.701 &
  {\color[HTML]{000000} 0.693} &
   &
  0.688 &
  0.684 &
  {\color[HTML]{000000} 0.690} \\
 &
  RF &
  0.770 &
  0.695 &
  {\color[HTML]{000000} 0.680} &
   &
  0.780 &
  0.687 &
  {\color[HTML]{000000} 0.688} \\
 &
  XGB &
  0.730 &
  0.703 &
  {\color[HTML]{0000FF} 0.702} &
   &
  0.705 &
  0.693 &
  {\color[HTML]{0000FF} \textbf{0.702}} \\
 &
  NN &
  0.767 &
  0.709 &
  {\color[HTML]{000000} 0.698} &
   &
  0.828 &
  0.674 &
  {\color[HTML]{000000} 0.668} \\ \hline
\end{tabular}
}
\end{table}


\begin{table*}[htbp]
\caption[]{AUC performance of champion/challenger models in production\protect\footnotemark on P2P dataset}
\label{tab:prod-table1}
\resizebox{\textwidth}{!} 
{
\begin{tabular}{ccllllllllllll}
\hline
\multirow{2}{*}{\textbf{\begin{tabular}[c]{@{}c@{}}Forecasting \\ method\end{tabular}}} &
  \multirow{2}{*}{\textbf{\begin{tabular}[c]{@{}c@{}}Challenger \\ models\end{tabular}}} &
  \multicolumn{12}{c}{\textbf{PROD monthly performance}} \\
 &
   &
  \multicolumn{1}{c}{\textbf{M1}} &
  \multicolumn{1}{c}{\textbf{M2}} &
  \multicolumn{1}{c}{\textbf{M3}} &
  \multicolumn{1}{c}{\textbf{M4}} &
  \multicolumn{1}{c}{\textbf{M5}} &
  \multicolumn{1}{c}{\textbf{M6}} &
  \multicolumn{1}{c}{\textbf{M7}} &
  \multicolumn{1}{c}{\textbf{M8}} &
  \multicolumn{1}{c}{\textbf{M9}} &
  \multicolumn{1}{c}{\textbf{M10}} &
  \multicolumn{1}{c}{\textbf{M11}} &
  \multicolumn{1}{c}{\textbf{M12}} \\ \hline
Fixed                & LR                   & 0.667 & 0.661 & 0.651 & 0.663 & 0.662 & 0.665 & 0.657 & 0.651 & 0.653 & 0.639 & 0.630 & 0.637  \\
Window               & RF                   & 0.653 & 0.640 & 0.633 & 0.643 & 0.636 & 0.639 & 0.631 & 0.628 & 0.624 & 0.614 & 0.601 & 0.609  \\
(Baseline                     & XGB         & {\color[HTML]{0000FF} 0.697} & 0.689 & 0.682 & 0.693 & 0.693 & 0.694 & 0.689 & 0.688 & 0.685 & 0.674 & 0.666 & 0.679  \\
Model)                     & NN             & 0.648 & 0.643 & 0.645 & 0.654 & 0.657 & 0.654 & 0.646 & 0.643 & 0.642 & 0.637 & 0.631 & 0.648 \\

\multicolumn{1}{l}{} & \multicolumn{1}{l}{} &        &         &        &        &        &        &         &         &         &         &         &         \\
Fixed              & LR                   & 0.667 & 0.662 & 0.651 & 0.663 & 0.661 & 0.664 & 0.658 & 0.650 & 0.653 & 0.674 & 0.668 & 0.677 \\
Origin               & RF                 & 0.653 & 0.642 & 0.635 & 0.648 & 0.645 & 0.648 & 0.645 & 0.654 & 0.65 & 0.650 & 0.645 & 0.656  \\
Recalibration        & XGB                & 0.697 & 0.691 & 0.689 & 0.707 & 0.708 & 0.707 & 0.705 & 0.698 & 0.703 & 0.694 & 0.687 & 0.701 \\
                     & NN                 & 0.648 & 0.639 & 0.629 & 0.643 & 0.638 & 0.652 & 0.660 & 0.634 & 0.659 & 0.679 & 0.687 & 0.685\\

\multicolumn{1}{l}{} & \multicolumn{1}{l}{} &        &         &        &        &        &        &         &         &         &         &         &         \\
Rolling              & LR                   & 0.667 & 0.662 & 0.651 & 0.663 & 0.659 & 0.663 & 0.656 & 0.648 & 0.651 & 0.669 & 0.666 & 0.673  \\
Window               & RF                   & 0.653 & 0.641 & 0.637 & 0.656 & 0.647 & 0.650 & 0.648 & 0.655 & 0.657 & 0.653 & 0.654 & 0.670  \\
                     & XGB                  & 0.697 & 0.690 & 0.698 & 0.703 & 0.703 & 0.710 & 0.703 & 0.701 & 0.695 & 0.689 & 0.689 & 0.702 \\
                     & NN                   & 0.648 & 0.636 & 0.616 & 0.628 & 0.631 & 0.631 & 0.646 & 0.613 & 0.633 & 0.648 & 0.663 & 0.670 \\

\multicolumn{1}{l}{} & \multicolumn{1}{l}{} &        &         &        &        &        &        &         &         &         &         &         &         \\
Champion               &  Model                   & $XGB^{FW}_{M1}$ & $XGB^{FO}_{M2}$ & $XGB^{RW}_{M3}$ & $XGB^{FO}_{M4}$ & $XGB^{FO}_{M4}$ & $XGB^{RW}_{M6}$ & $XGB^{RW}_{M6}$ & $XGB^{RW}_{M6}$ & $XGB^{FO}_{M9}$ & $XGB^{FO}_{M9}$ & $XGB^{RW}_{M11}$ & $XGB^{RW}_{M11}$  \\

Model with               & AUC                   & {\color[HTML]{0000FF} 0.697} & {\color[HTML]{0000FF} 0.691} & {\color[HTML]{0000FF} 0.699} & {\color[HTML]{0000FF} 0.707} & {\color[HTML]{0000FF} 0.707} & {\color[HTML]{0000FF} 0.710} & {\color[HTML]{0000FF} 0.706} & {\color[HTML]{0000FF} 0.704} & {\color[HTML]{0000FF} 0.703} & {\color[HTML]{0000FF} 0.695} & {\color[HTML]{0000FF} 0.689} & {\color[HTML]{0000FF} 0.701}  \\

our Algorithm            & Performance ($\Delta$)           &  &  & &  & &  &  & & & & &  \\
                     & over baseline                  & 0.0\%	& 0.2\%	& 1.7\%	& 1.4\%	& 1.4\%	& 1.6\%	& 1.7\%	& 1.6\%	& 1.8\%	& 2.1\%	& 2.3\%	& 2.2\% \\ \hline

\end{tabular}
}
\end{table*}
\footnotetext{Bolded values represent the AUC of the champion model; values in red represent the performance improvement of rolling origin recalibration or rolling window compared to the fixed origin method.}


\begin{table*}[htbp]
\caption[]{AUC performance of champion/challenger models in production on SBA dataset}
\label{tab:prod-table2}
\resizebox{\textwidth}{!} 
{
\begin{tabular}{ccllllllllllll}
\hline
\multirow{2}{*}{\textbf{\begin{tabular}[c]{@{}c@{}}Forecasting \\ method\end{tabular}}} &
  \multirow{2}{*}{\textbf{\begin{tabular}[c]{@{}c@{}}Challenger \\ models\end{tabular}}} &
  \multicolumn{12}{c}{\textbf{PROD quarterly performance}} \\
 &
   &
  \multicolumn{1}{c}{\textbf{Q1}} &
  \multicolumn{1}{c}{\textbf{Q2}} &
  \multicolumn{1}{c}{\textbf{Q3}} &
  \multicolumn{1}{c}{\textbf{Q4}} &
  \multicolumn{1}{c}{\textbf{Q5}} &
  \multicolumn{1}{c}{\textbf{Q6}} &
  \multicolumn{1}{c}{\textbf{Q7}} &
  \multicolumn{1}{c}{\textbf{Q8}} &
  \multicolumn{1}{c}{\textbf{Q9}} &
  \multicolumn{1}{c}{\textbf{Q10}} &
  \multicolumn{1}{c}{\textbf{Q11}} &
  \multicolumn{1}{c}{\textbf{Q12}} \\ \hline
Fixed                & LR                   & 0.776 & 0.774 & 0.758 & 0.762 & 0.753 & 0.763 & 0.704 & 0.724 & 0.724 & 0.676 & 0.670 & 0.715  \\
Window               & RF                   & 0.878 & 0.869 & 0.855 & 0.848 & 0.840 & 0.850 & 0.783 & 0.804 & 0.822 & 0.779 & 0.784 & 0.826  \\
(Baseline                     & XGB         & {\color[HTML]{0000FF} 0.938} & 0.933 & 0.926 & 0.907 & 0.917 & 0.901 & 0.882 & 0.901 & 0.914 & 0.885 & 0.858 & 0.889  \\
Model)                     & NN             & 0.782 & 0.788 & 0.792 & 0.779 & 0.735 & 0.738 & 0.701 & 0.731 & 0.736 & 0.719 & 0.701 & 0.702 \\

\multicolumn{1}{l}{} & \multicolumn{1}{l}{} &        &         &        &        &        &        &         &         &         &         &         &         \\
Fixed              & LR                   & 0.776 & 0.775	& 0.756	& 0.765	& 0.777	& 0.769	& 0.709	& 0.733	& 0.734	& 0.696	& 0.675	& 0.728  \\
Origin               & RF                 & 0.878 & 0.880	& 0.880	& 0.863	& 0.905	& 0.886	& 0.884	& 0.907	& 0.923	& 0.902	& 0.872	& 0.916  \\
Recalibration        & XGB                & 0.938 & 0.943	& 0.942	& 0.928	& 0.930	& 0.928	& 0.924	& 0.941	& 0.935	& 0.927	& 0.919	& 0.934  \\
                     & NN                 & 0.772 & 0.775	& 0.764	& 0.783	& 0.774	& 0.787	& 0.820	& 0.865	& 0.873	& 0.860	& 0.860	& 0.901  \\

\multicolumn{1}{l}{} & \multicolumn{1}{l}{} &        &         &        &        &        &        &         &         &         &         &         &         \\
Rolling              & LR                   & 0.776	& 0.774	& 0.759	& 0.765	& 0.783	& 0.776	& 0.721	& 0.751	& 0.742	& 0.706	& 0.688	& 0.744  \\
Window               & RF                   & 0.878	& 0.882	& 0.888	& 0.882	& 0.865	& 0.893	& 0.831	& 0.913	& 0.931	& 0.914	& 0.923	& 0.942  \\
                     & XGB                  & 0.938	& 0.938	& 0.949	& 0.931	& 0.937	& 0.950	& 0.944	& 0.955	& 0.948	& 0.951	& 0.939	& 0.905 \\
                     & NN                   & 0.789	& 0.784	& 0.787	& 0.772	& 0.792	& 0.697	& 0.839	& 0.834	& 0.861	& 0.851	& 0.873	& 0.898 \\

\multicolumn{1}{l}{} & \multicolumn{1}{l}{} &        &         &        &        &        &        &         &         &         &         &         &         \\
Champion               &  Model                   & $XGB^{FW}_{M1}$ & $XGB^{FO}_{M2}$ & $XGB^{RW}_{M3}$ & $XGB^{RW}_{M3}$ & $XGB^{RW}_{M3}$ & $XGB^{RW}_{M3}$ & $XGB^{RW}_{M7}$ & $XGB^{RW}_{M8}$ & $XGB^{RW}_{M9}$ & $XGB^{RW}_{M10}$ & $XGB^{RW}_{M11}$ & $XGB^{RW}_{M11}$  \\

Model with               & AUC                   & {\color[HTML]{0000FF} 0.938} & {\color[HTML]{0000FF} 0.943} & {\color[HTML]{0000FF} 0.949} & {\color[HTML]{0000FF} 0.938} & {\color[HTML]{0000FF} 0.943} & {\color[HTML]{0000FF} 0.950} & {\color[HTML]{0000FF} 0.944} & {\color[HTML]{0000FF} 0.955} & {\color[HTML]{0000FF} 0.948} & {\color[HTML]{0000FF} 0.951} & {\color[HTML]{0000FF} 0.939} & {\color[HTML]{0000FF} 0.948}  \\

our Algorithm            & Performance ($\Delta$)           &  &  & &  & &  &  & & & & &  \\
                     & over baseline                  & 0.0\%	& 1.0\%	& 2.3\%	& 3.1\%	& 2.5\%	& 4.9\%	& 6.2\%	& 5.3\%	& 3.4\%	& 6.6\%	& 8.2\%	& 5.9\% \\ \hline

\end{tabular}
}
\end{table*}


\begin{table*}[htbp]
\caption[]{AUC performance of champion/challenger models in production on AutoL dataset}
\label{tab:prod-table3}
\resizebox{\textwidth}{!} 
{
\begin{tabular}{ccllllllllllll}
\hline
\multirow{2}{*}{\textbf{\begin{tabular}[c]{@{}c@{}}Forecasting \\ method\end{tabular}}} &
  \multirow{2}{*}{\textbf{\begin{tabular}[c]{@{}c@{}}Challenger \\ models\end{tabular}}} &
  \multicolumn{12}{c}{\textbf{PROD monthly performance}} \\
 &
   &
  \multicolumn{1}{c}{\textbf{M1}} &
  \multicolumn{1}{c}{\textbf{M2}} &
  \multicolumn{1}{c}{\textbf{M3}} &
  \multicolumn{1}{c}{\textbf{M4}} &
  \multicolumn{1}{c}{\textbf{M5}} &
  \multicolumn{1}{c}{\textbf{M6}} &
  \multicolumn{1}{c}{\textbf{M7}} &
  \multicolumn{1}{c}{\textbf{M8}} &
  \multicolumn{1}{c}{\textbf{M9}} &
  \multicolumn{1}{c}{\textbf{M10}} &
  \multicolumn{1}{c}{\textbf{M11}} &
  \multicolumn{1}{c}{\textbf{M12}} \\ \hline
Fixed                & LR                   & 0.633	& 0.622	& 0.627	& 0.606	& 0.625	& 0.606	& 0.583	& 0.564	& 0.547	& 0.528	& 0.543	& 0.539  \\
Window               & RF                   & {\color[HTML]{0000FF} 0.657}	& 0.671	& 0.655	& 0.648	& 0.654	& 0.623	& 0.592	& 0.602	& 0.585	& 0.565	& 0.600	& 0.599  \\
(Baseline            & XGB         & 0.633	& 0.650	& 0.631	& 0.631	& 0.611	& 0.622	& 0.610	& 0.600	& 0.568	& 0.557	& 0.586	& 0.574  \\
Model)               & NN             & 0.650	& 0.642	& 0.637	& 0.617	& 0.634	& 0.611	& 0.584	& 0.595	& 0.569	& 0.551	& 0.571	& 0.586 \\

\multicolumn{1}{l}{} & \multicolumn{1}{l}{} &        &         &        &        &        &        &         &         &         &         &         &         \\
Fixed                & LR                   & 0.633	& 0.619	& 0.625	& 0.623	& 0.626	& 0.611	& 0.608	& 0.587	& 0.568	& 0.537	& 0.556	& 0.543  \\
Origin               & RF                 & 0.657	& 0.672	& 0.648	& 0.642	& 0.654	& 0.624	& 0.594	& 0.608	& 0.593	& 0.578	& 0.612	& 0.609  \\
Recalibration        & XGB                & 0.633	& 0.680	& 0.654	& 0.634	& 0.648	& 0.645	& 0.616	& 0.615	& 0.596	& 0.588	& 0.625	& 0.610  \\
                     & NN                 & 0.660	& 0.641	& 0.625	& 0.629	& 0.629	& 0.603	& 0.584	& 0.588	& 0.570	& 0.572	& 0.583	& 0.597  \\

\multicolumn{1}{l}{} & \multicolumn{1}{l}{} &        &         &        &        &        &        &         &         &         &         &         &         \\
Rolling              & LR                   & 0.633	& 0.621	& 0.617	& 0.594	& 0.624	& 0.611	& 0.593	& 0.583	& 0.568	& 0.537	& 0.566	& 0.540  \\
Window               & RF                   & 0.657	& 0.669	& 0.650	& 0.639	& 0.653	& 0.619	& 0.590	& 0.602	& 0.598	& 0.591	& 0.619	& 0.615  \\
                     & XGB                  & 0.633	& 0.688	& 0.663	& 0.648	& 0.625	& 0.635	& 0.606	& 0.610	& 0.595	& 0.608	& 0.626	& 0.628 \\
                     & NN                   & 0.649	& 0.648	& 0.625	& 0.619	& 0.627	& 0.590	& 0.585	& 0.585	& 0.575	& 0.567	& 0.597	& 0.599 \\

\multicolumn{1}{l}{} & \multicolumn{1}{l}{} &        &         &        &        &        &        &         &         &         &         &         &         \\
Champion             &  Model                   & $RF^{FW}_{M1}$ & $XGB^{RW}_{M2}$ & $XGB^{RW}_{M3}$ & $XGB^{RW}_{M4}$ & $XGB^{RW}_{M4}$ & $XGB^{FO}_{M6}$ & $XGB^{FO}_{M6}$ & $XGB^{FO}_{M8}$ & $XGB^{FO}_{M9}$ & $XGB^{FO}_{M10}$ & $XGB^{RW}_{M11}$ & $XGB^{RW}_{M12}$  \\

Model with           & AUC                   & {\color[HTML]{0000FF} 0.657} & {\color[HTML]{0000FF} 0.688} & {\color[HTML]{0000FF} 0.663} & {\color[HTML]{0000FF} 0.648} & {\color[HTML]{0000FF} 0.657} & {\color[HTML]{0000FF} 0.645} & {\color[HTML]{0000FF} 0.615} & {\color[HTML]{0000FF} 0.615} & {\color[HTML]{0000FF} 0.596} & {\color[HTML]{0000FF} 0.588} & {\color[HTML]{0000FF} 0.626} & {\color[HTML]{0000FF} 0.628}  \\

our Algorithm         & Performance ($\Delta$)           &  &  & &  & &  &  & & & & &  \\
                     & over baseline                  & 0.0\%	& 1.7\%	& 0.8\%	& 0.0\%	& 0.3\%	& 2.2\%	& 2.3\%	& 1.3\%	& 1.1\%	& 2.3\%	& 2.6\%	& 3.0\% \\ \hline

\end{tabular}
}
\end{table*}

We further delved into understanding why model performance declines over time after deployment and how PDx sustains high predictive accuracy in the production environment. A key limitation of the fixed window baseline model in production is its inability to self-update in response to the latest customer attributes, failing to address data drift, or adapting to changes in repayment behavior, known as concept drift. Our analysis of model iterations before production revealed that each new model version adjusted variable weights and importance, enabling these updated models to capture shifts in the behavior of incoming loan applicants. Table \ref{tab:shap} illustrates the SHAP value differences between the baseline $XGB^{FW}$ model used in the first production month and the $XGB^{RW}_{M11}$, a rolling window PDx champion model trained just before the 11th production month and applied in the 11th and 12th months. Notably, there are significant differences not only in the number of variables used but also in their ranking and influence, highlighting the adaptive nature of the PDx approach.


\begin{table*}[htbp]
    \centering
    \begin{tabular}{c c}
        \hline
        Month 1 (Baseline FW XGB) & Month 12 (PDx - $XGB^{RW}_{M11}$)  \\ \hline
        \includegraphics[width=0.4\linewidth]{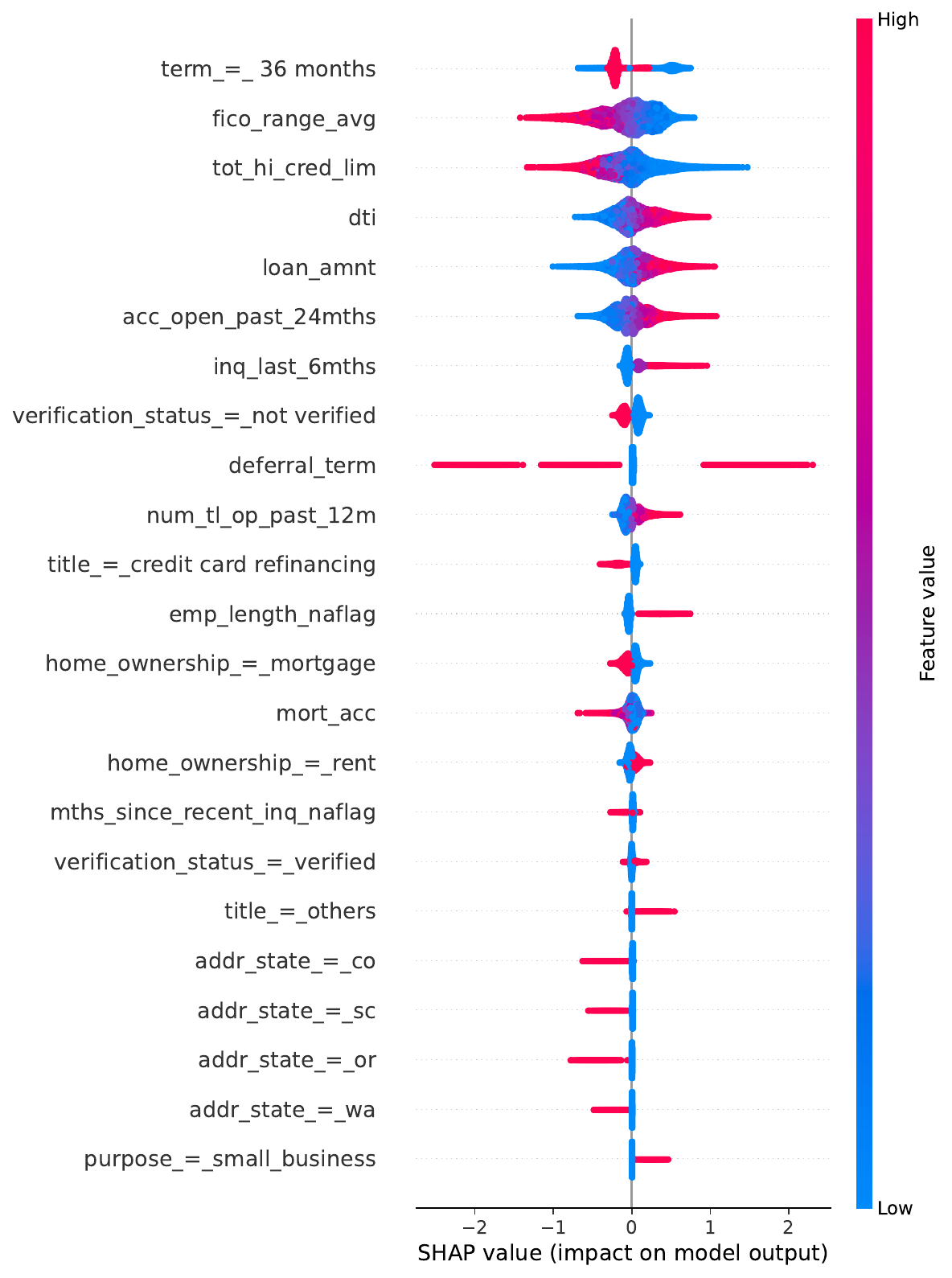} & \includegraphics[width=0.4\linewidth]{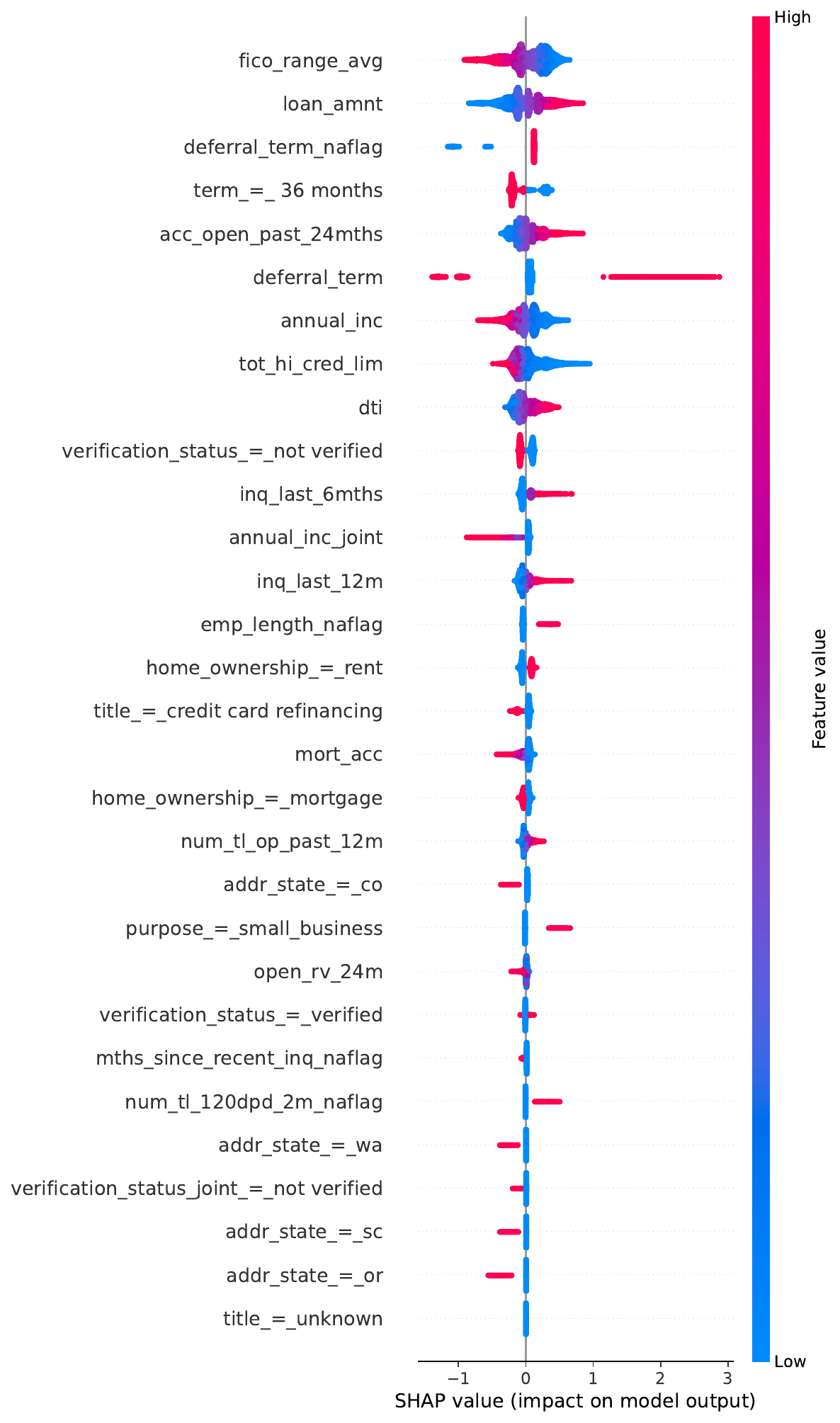} \\ \hline
    \end{tabular}
    \caption{Comparison of SHAP plots for first and 12th month (P2P experiment)}
    \label{tab:shap}
\end{table*}


\section{Discussion}
\label{sec:discussion}
\subsection{Methodological contributions}

Our proposed algorithm, PDx advances credit default forecasting by integrating continuous learning and MLOps, ensuring sustained predictive performance of machine learning models in production environments. Unlike traditional approaches, PDx is designed for adaptive decision-making, enabling models to remain robust and up-to-date. A key innovation of PDx lies in its ability to enhance model stability and preemptive failure management by systematically evaluating predictive power and variable interpretability on the most recent data, ensuring a balanced and transparent approach to credit risk assessment.
Methodological contribution of our work can be attributed to three critical challenges it overcomes in existing ML-based forecasting methodologies for business applications:

\begin{enumerate}
    \item Model Performance Stress Testing: Traditional methods rely on out-of-sample validation or cross-validation within the same data window, which is insufficient for evaluating real-world performance. PDx incorporates \textbf{out-of-time (OTV) validation} to test models on \textbf{future sample data}, providing a more reliable measure of predictive improvement.
    
    \item Continuous Model Retraining Strategies: PDx enables two retraining mechanisms—\textbf{fixed-origin recalibration (data append)}, which expands the training dataset with new observations, and \textbf{rolling-window updates (data replace)}, which prioritizes recency by replacing older data. The choice between these strategies is not binary; an optimal combination depends on the underlying data characteristics and business needs.
    
    \item Mitigating Value Erosion through Continuous Retraining and Deployment: PDx ensures that models are continuously updated to prevent performance degradation, demonstrating that sustained predictive accuracy is not baseline algorithm-dependent but rather a function of model adaptability.
\end{enumerate}

Additionally, we demonstrate how cloud-based architectures facilitate seamless model deployment, supporting end-to-end automation in digital lending within a fully operational ML lifecycle. Beyond credit risk forecasting, PDx framework is highly generalizable and can be applied to churn prediction, fraud detection, predictive maintenance, and other time-sensitive forecasting applications. While this paper focuses on a classification problem, the framework can be extended to other machine learning tasks, including regression, time series forecasting, and survival analysis modeling.

\subsection{Practical and managerial contributions}

According to a 2021 report by Fair Isaac Corporation (FICO) and Corinium, many companies struggle with AI adoption, with 65\% unable to explain model decisions, 73\% lacking executive support for AI ethics, and only 20\% actively monitoring models in production \cite{baesens2023boosting}. This lack of oversight is particularly critical in digital lending, where misclassifying a bad loan incurs significantly higher costs than earnings from good loans \cite{lyocsa2022default}. Studies suggest that one defaulted loan may require the revenue from five to ten performing loans to offset losses, depending on loan terms and recovery rates. This imbalance leads to a significant increase in non-performing assets (NPAs), creating liquidity challenges for lenders and underscoring the need for robust credit risk modeling and continuous model monitoring

With PDx, we address key challenges in credit risk forecasting by automating model updates, shifting from periodic manual adjustments to a seamless, continuous improvement cycle. This enhances operational efficiency and allows institutions to reallocate resources from manual credit and model risk management to strategic decision-making while maintaining model accuracy in an evolving financial landscape. As digital lending accelerates, keeping models up to date is critical, and PDx serves as a key enabler. By leveraging fixed-origin recalibration and rolling-window updates, PDx ensures alignment with changing market trends and customer behaviors, reducing the risk of model obsolescence. This adaptive approach strengthens risk management while providing a competitive advantage, enabling financial institutions to make more informed and resilient credit decisions. 

The practical and managerial impact of PDx lies in its ability to drive long-term incremental value for business by enabling more accurate and adaptive credit decisions. Across all three experiments, PDx consistently outperformed traditional fixed-window models, regardless of the underlying algorithm. Beyond technical KPIs such as AUC, KS, and F1 Score, PDx demonstrated its effectiveness in identifying a higher proportion of defaulters within the top three deciles, directly improving risk mitigation. Additionally, PDx enhances regulatory compliance by automating model risk management reports, providing insights into model health, interpretability, and fairness, thereby helping financial institutions meet evolving regulatory standards efficiently.

\subsection{Limitations and further work}
This research has several limitations, primarily due to its reliance on a limited dataset, which may restrict the comprehensive evaluation of model performance. Additionally, the dataset consists exclusively of approved loan applications, omitting rejected cases and their performance on off-us (loan granted by other lenders) loan data, potentially limiting the model's robustness. One key factor contributing to performance degradation is population behavior shifts, which could be further examined by analyzing the Population Stability Index (PSI) over out-of-time validation (OTV) and production months. However, due to data constraints, this analysis could not be conducted in the current study and remains an area for future exploration using additional data, particularly from the rejected loan base. Furthermore, while our framework can assess fairness, the absence of protected attributes such as gender and race prevents a comprehensive fairness evaluation in credit scoring and the impact of continuous model development and monitoring.

For future research, incorporating rejected loan applications and off-us loan data would enable a more comprehensive performance assessment. Additionally, evaluating PDx’s role in fairness assessment within a production environment would provide more valuable insights. Further exploration of techniques to address class imbalance and enhance fairness in model predictions would contribute to more equitable lending practices. Beyond credit risk assessment, continuous learning and integration could be extended to other areas of consumer risk modeling, such as Exposure at Default (EAD) and Loss Given Default (LGD). Moreover, assessing the impact of more complex ensemble algorithms, such as stacking models, could refine baseline model selection. Finally, investigating the role of MLOps in broader banking applications, including marketing, operations, and financial management, could unlock additional opportunities for automation and optimization. We believe addressing these limitations and exploring future research directions can further enhance the credit risk forecasting models for digital lending applications.

\section{Conclusion}
\label{sec:conclusion}
In this paper, we introduced PDx, a novel framework that integrates MLOps to enhance credit risk forecasting in digital lending. Our empirical findings demonstrate PDx's ability to improve predictive accuracy, maintain model robustness, and enable real-time adaptation to evolving market conditions. By embedding continuous learning within the model lifecycle, PDx represents a significant advancement in financial technology, addressing key challenges in model degradation and adaptability. Beyond probability of default (PD) modeling, this framework provides a scalable foundation for refining broader credit risk assessment models.

Our research underscores the importance of adaptability and precision in financial risk management by leveraging automated model updates and advanced machine learning operations. The implications extend beyond credit scoring, highlighting potential applications across financial services and risk assessment domains. Moving forward, further exploration of machine learning innovations in financial decision-making can drive greater efficiency, improved customer outcomes, and higher industry standards. These insights pave the way for future advancements in risk modeling, regulatory compliance, and AI-driven financial decision systems.

\section{Declaration of Competing Interest}
The authors declare no conflicts of interest in this study. They bear sole responsibility for both the content and the composition of this manuscript.

\section{Data Availability Statement}
\noindent The data that support the findings of this study are available from the corresponding author upon reasonable request.

\appendix


\section{Machine Learning Models}

\subsection{Logistic Regression (LR)}
LR is one of the most widely used models in traditional credit risk scoring engines. It is a simple yet interpretable classical linear model that gives the probabilities of whether a customer will default:
    \begin{align*}
        P\left(\mathbf{y_i} = 1 | \mathbf{x_i}, \mathbf{w}\right) &= \sigma\left(\mathbf{w}^{\intercal} \mathbf{x_i}\right) \\
        &= \frac{1}{1 + \exp\left(\mathbf{w}^{\intercal} \mathbf{x_i}\right)}
    \end{align*}
    Here, \( \mathbf{x_i} \) denotes the feature vector for the \( i \)-th instance, and \( \mathbf{w} \) represents the weight vector. The logistic function transforms the linear combination \( \mathbf{w}^{\intercal} \mathbf{x_i} \) into a probability between 0 and 1.

    In this paper, we have also incorporated shrinkage methods into Logistic Regression to mitigate overfitting, which is particularly crucial in high-dimensional settings where the number of features \( N \) is large relative to the number of instances. Initially, \( l_1 \) regularization, also known as Lasso, is employed for feature selection. This regularization technique encourages sparsity in the model, effectively performing feature selection by driving certain coefficients to zero. The mathematical formulation for Lasso is:
    \begin{align*}
    \text{minimize} \quad \frac{1}{2M} \sum_{i=1}^{M} \left(y_i - \mathbf{w}^{\intercal} \mathbf{x_i}\right)^2 + \lambda_1 \|\mathbf{w}\|_1 
    \end{align*}
    Subsequently, \( l_2 \) regularization, or Ridge, is applied to further refine the model performance. Ridge regularization helps in shrinking the coefficients, thereby reducing model complexity and preventing overfitting. Its formulation is:
    \begin{align*}
    \text{minimize} \quad \frac{1}{2M} \sum_{i=1}^{M} \left(y_i - \mathbf{w}^{\intercal} \mathbf{x_i}\right)^2 + \lambda_2 \|\mathbf{w}\|_2^2 
    \end{align*}
    where \( M \) represents the number of instances, and \( \lambda_1 \) and \( \lambda_2 \) are the regularization parameters for Lasso and Ridge, respectively. Logistic Regression excels in interpretability, simplicity, and computational efficiency. However, it is less effective in handling missing values, outliers, class imbalance, and multicollinearity. Additionally, it cannot capture complex relationships and variable interactions as effectively as XGBoost, Random Forest, and NN-MLP.LR demonstrates consistent and stable prediction capabilities, as evidenced in Table \ref{tab:dev-table}, which shows that LR exhibits the smallest performance gap among the Development (DEV), In-Time Validation (ITV), and Out-of-Time Validation (OTV) datasets. \\


\subsection{Random Forest (RF)}
RF \cite{breiman2001random} is an ensemble method that uses the bootstrap aggregating (bagging) approach whereby multiple decision trees are fitted on subsamples of the dataset, and the output scores are combined to form a final prediction. It is robust to multicollinearity in features, outliers, and overfitting and has proven to be one of the best classifiers for default prediction \cite{dastile2020statistical}. For dataset \( D_i \) with \( M \) observations and \( N \) features where each observation is represented as \( \mathbf{x}_i = (x_{i1}, x_{i2}, \ldots, x_{iN}) \) with a corresponding label \( y_i \), RF builds an ensemble of \( K \) decision trees, denoted as \( \{T_1, T_2, \ldots, T_K\} \). For each tree \( T_k \), a bootstrap sample \( D_{i,k}^* \) is drawn from \( D_i \). For tree \( T_k \), a subset of features \( F_k \subseteq \{1, 2, \ldots, N\} \) with \( |F_k| = p \) is randomly selected. The tree is grown by recursively partitioning \( D_{i,k}^* \) based on feature splits that maximize the decrease in a criterion, typically Gini impurity:
    \[ G(S) = 1 - \sum_{c \in \text{classes}} (p_c)^2 \]
    
    Each split \( s \) on feature \( j \) in node \( t \) is chosen to maximize the impurity decrease:
    \begin{align*}
    \Delta G(t, s) &= G(t) - \left( \frac{|D_{t, \text{left}}(s)|}{|D_t|} G(D_{t, \text{left}}(s)) \right. \\
    &\qquad \left. + \frac{|D_{t, \text{right}}(s)|}{|D_t|} G(D_{t, \text{right}}(s)) \right)
    \end{align*}

     Post training, feature importance for feature \( j \) is calculated as the average decrease in Gini impurity:
    \[ \text{Importance}(j) = \frac{1}{K} \sum_{k=1}^{K} \sum_{t \in T_k} \mathbf{1}(j \in t) \cdot \Delta G(t, s_t) \]
    where \( \mathbf{1}(j \in t) \) is an indicator function that is 1 if feature \( j \) is used in split \( s_t \) at node \( t \) in tree \( T_k \). An initial fit of the model is performed for feature selection according to the feature importance scores, followed by a final fit of the model for evaluation.\\


\subsection{XGBoost (XGB)}

XGB\cite{chen2016xgboost} has been extensively used in various ML problem-solving competitions, including and often appearing as one of the top solutions. It is a gradient-boosting method that creates a strong model from an ensemble of weak learners. For a dataset \( D_i \) with \( M \) observation and each observation having N number of features, denoted as \( X_i = \{x_{i1}, x_{i2}, \ldots, x_{iN}\} \) the model aims to predict a target variable \( y_i \) (e.g., credit scoring \citep{chen2019application,yotsawat2021improved}, ) for each instance. XGBoost builds a series of \( K \) decision trees (weak learners) in a sequential manner, where each tree \( T_k \) attempts to minimize the residuals of its predecessor models iteratively. This sequential approach, where the prediction for instance \( \mathbf{x}_i \) at iteration \( k \) is updated by 
\[ \hat{y}_i^{(k)} = \hat{y}_i^{(k-1)} + \eta \cdot f_k(\mathbf{x}_i) 
\] with \( f_k \) as the \( k \)-th tree and \( \eta \) as the learning rate, allows XGB to handle datasets with complex relationships effectively. To minimize loss, we have used the below objective functions:  
\[\text{Obj} = \sum_{i=1}^{M} \ell(y_i, \hat{y}_i^{(K)}) + \sum_{k=1}^{K} \Omega(f_k) \] where \( \ell(y_i, \hat{y}_i^{(K)}) \) is a loss function which in our case is a logistic loss and \( \Omega(f_k) \) is a regularization term defined as:

\[ \Omega(f_k) = \gamma T + \frac{1}{2} \lambda \sum_{j=1}^{T} w_j^2 \] where \( T \) is the number of leaves, \( w_j \) is the score on leaf \( j \), and \( \gamma, \lambda \) are regularization parameters. Similar to Random Forest, XGB also offers feature importance, and it is estimated based on the gain (improvement in accuracy) from splits on feature \( j \) across all trees: \[ \text{Importance}(j) = \sum_{k=1}^{K} \sum_{t \in \text{splits}_k} \Delta \text{Gain}(j, t) \]  where \( \text{splits}_k \) are the splits in tree \( T_k \) and \( \Delta \text{Gain}(j, t) \) is the gain from split \( t \) on feature \( j \). XGB can effectively handle missing values, class imbalance \citep{mushava2022novel}, and outliers in dataset \( D_i \). It can also be easily interpreted \citep{biecek2021enabling,liu2023credit} using bi-variate, partial dependency, and SHAP plots. Similar to RF, an initial fit is performed for feature selection, followed by a final fit for evaluation. \\

\subsection{Neural Network (NN)}
As its name suggests, NN is inspired by the brain anatomy, which comprises interconnected nodes, also known as neurons, that propagate information across the network. While many different forms of NN architectures exist, i.e., recurrent neural networks, convolutional neural networks, and transformers, we have selected a basic multi-layer perceptron (MLP) \cite{pinkus1999approximation} architecture for our experiment, which has proven to be one of the best baseline models \citep{papouskova2019two,abellan2017comparative}.
    
    The MLP model is structured as follows:

\begin{enumerate}
    \item \textbf{Input Layer}: It has \( N \) nodes corresponding to the \( N \) features of the dataset \( D_i \), denoted as \( \mathbf{X}_i = [x_{i1}, x_{i2}, \ldots, x_{iN}]^\top \) for each instance \( i \).
    \item \textbf{Hidden Layers}: There are three hidden layers, each comprising 150 units. The output of each hidden layer \( j \) is represented as \( \mathbf{H}_j = \text{ReLU}(\mathbf{W}_j \mathbf{H}_{j-1} + \mathbf{b}_j) \), where \( \mathbf{W}_j \) and \( \mathbf{b}_j \) are the weight matrix and bias vector for layer \( j \), respectively, and \( \mathbf{H}_0 = \mathbf{X}_i \). A dropout rate of 10\% is applied to the first and third hidden layers for regularization, denoted by \( \text{Dropout}(\mathbf{H}_j, 0.10) \).
    \item \textbf{Output Layer}: The network culminates in an output layer with a single unit. The final output \( \hat{y}_i \) is obtained using a sigmoid activation function: \( \hat{y}_i = \sigma(\mathbf{W}_O \mathbf{H}_3 + b_O) \), ensuring that the output values are constrained to the range (0, 1).
    \item \textbf{Parameter Count}: The model encompasses 79,051 parameters derived from the sum of the weights and biases of each layer.
    \item \textbf{Loss Function and Optimization}: The NN is trained to minimize the binary cross-entropy loss function, \( \mathcal{L}(\mathbf{y}, \hat{\mathbf{y}}) = -\frac{1}{M}\sum_{i=1}^{M}[y_i\log(\hat{y}_i) + (1-y_i)\log(1-\hat{y}_i)] \), where \( M \) is the number of instances and \( y_i \) is the true label. Optimization is performed using the Adam optimizer, a variant of stochastic gradient descent.
\end{enumerate}
    
    The NN-MLP model effectively learns to perform credit scoring tasks by adjusting its parameters during training to minimize the loss function, leveraging the ReLU activation for nonlinearity and dropout for regularization. NN MLP offers high flexibility and potential for capturing complex patterns but comes with challenges in interpretability, overfitting, and computational demands against LR, XGB, and RF. This observation is evident in Table \ref{tab:dev-table}, which illustrates that, during model development, the NN-MLP model exhibits the largest performance gap between the Development (DEV), In-Time Validation (ITV), and Out-of-Time Validation (OTV) datasets.\\

\bibliographystyle{apalike}  
\bibliography{aipsamp.bib}

\end{document}